\documentclass[10pt,twocolumn,letterpaper]{article}

\usepackage{iccv}
\usepackage{times}
\usepackage{epsfig}
\usepackage{graphicx}
\usepackage{amsmath}
\usepackage{amssymb}

\usepackage{microtype}
\usepackage{todonotes}
\usepackage{tikz}
\usepackage{pgfplots}
\usepackage{xcolor}
\usepackage{float}
\usepackage{subfloat}
\usepackage{bm}
\usepackage{multirow}
\usepackage{colortbl}
\usepackage{adjustbox}
\usepackage[caption=false]{subfig}
\usepackage{enumitem}

\def\abbrmethod{SOAR}
\def\reconmod{AdRecon}
\def\sclsmod{AdaScls}

\definecolor{mybluecolor}{RGB}{0, 168, 234}
\definecolor{myredcolor}{RGB}{254, 0, 19}
\newcommand{\myblue}[1]{\textcolor{mybluecolor}{#1}}
\newcommand{\myred}[1]{\textcolor{myredcolor}{#1}}
\definecolor{mygray}{gray}{0.9}  
\definecolor{lightblue}{RGB}{194, 214, 236}
\definecolor{lightgreen}{RGB}{202, 223, 184}

\usepackage[pagebackref=true,breaklinks=true,letterpaper=true,colorlinks,bookmarks=false]{hyperref}

\usepackage[capitalize]{cleveref}
\crefname{section}{Sec.}{Secs.}
\Crefname{section}{Section}{Sections}
\Crefname{table}{Table}{Tables}
\crefname{table}{Tab.}{Tabs.}

\iccvfinalcopy 

\def\iccvPaperID{5263} 

\ificcvfinal\fi

\begin{document}

\title{SOAR: Scene-debiasing Open-set Action Recognition}

\author{Yuanhao Zhai$^1$\thanks{Work done during an internship at Wormpex AI Research.} ,
Ziyi Liu$^2$\thanks{Corresponding author.} ,
Zhenyu Wu$^2$,
Yi Wu$^2$,
Chunluan Zhou$^2$,
David Doermann$^1$, \\
Junsong Yuan$^1$,
Gang Hua$^2$ \\
$^1$University at Buffalo ~~ $^2$Wormpex AI Research \\
\footnotesize \texttt{\{yzhai6, doermann, jsyuan\}@buffalo.edu},~\texttt{\{wuzhenyusjtu, ywu.china, czhou002, ganghua\}@gmail.com}
}

\maketitle
\ificcvfinal\thispagestyle{empty}\fi


\begin{abstract}
Deep learning models have a risk of utilizing spurious clues to make predictions,
such as recognizing actions based on the background scene.
This issue can severely degrade the open-set action recognition performance when
the testing samples have different scene distributions from the training
samples.
To mitigate this problem, we propose a novel method, called Scene-debiasing Open-set Action Recognition~(\abbrmethod{}), which features an adversarial scene reconstruction
module and an adaptive adversarial scene classification module.
The former prevents the decoder from reconstructing the video background given
video features, and thus helps reduce the background information in feature
learning.
The latter aims to confuse scene type classification given video features,
with a specific emphasis on the action foreground, and
helps to learn scene-invariant information.
In addition, we design an experiment to quantify the scene bias.
The results indicate that the current open-set action recognizers are biased toward the
scene, and our proposed \abbrmethod{} method better mitigates such bias.
Furthermore, our extensive experiments demonstrate that our method outperforms state-of-the-art
methods, and the ablation studies confirm the effectiveness of our proposed
modules.
\end{abstract}

\section{Introduction}
\label{sec:introduction}

Recent years have witnessed significant progress in action
recognition~\cite{carreira2017quo,wang2018temporal,wang2018non,lin2019tsm,feichtenhofer2019slowfast,yang2020temporal,yu2019temporal,guo2022uncertainty,wu2022txvad}.
Yet, most works follow a closed-set paradigm, where both training and testing
videos belong to a set of pre-defined action categories.
This limits their application as the real world is naturally open with unknown
actions.
Open-set recognition is proposed to identify unknown samples from known ones
while maintaining classification performance on known
samples~\cite{scheirer2014probability,jain2014multi,bendale2015towards}.
It is challenging due to missing knowledge of the unknown world.
Moreover, deep models are found to rely on spurious information to make
predictions, \eg{}, classify images using local
textures~\cite{geirhos2018imagenet,mo2021object} and recognize actions using
background scene~\cite{li2018resound,choi2019can}.
This not only hurts the performance under the closed-set setting when training
and testing sets are not independent and identically distributed, but also
severely degrades the open-set recognition performance, as the distribution of
the open-set testing set is unknown.

\begin{figure}[t!]
    \includegraphics[width=\linewidth]{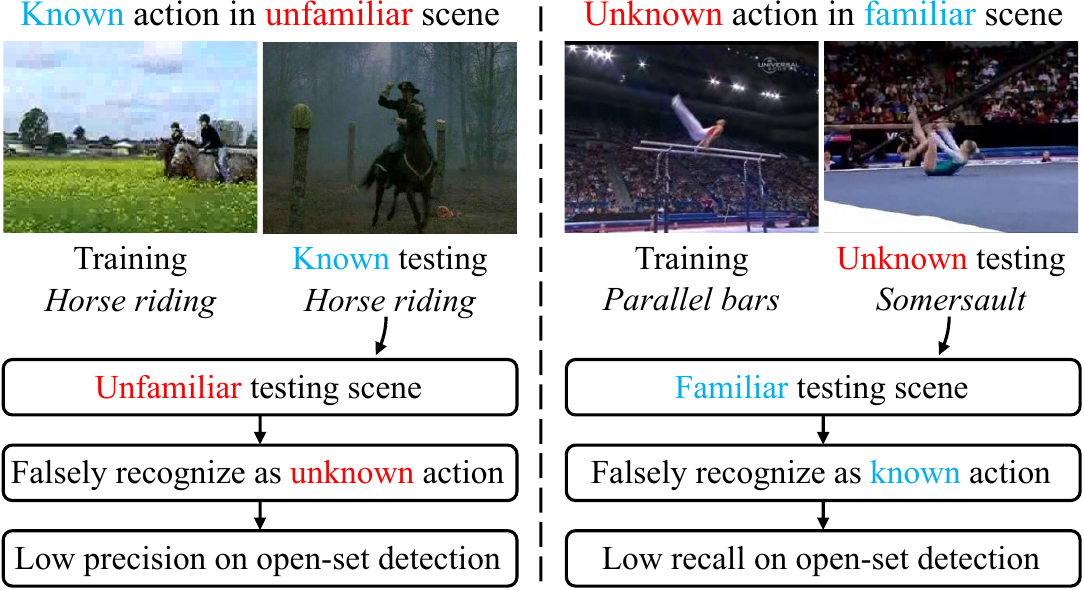}
    \caption{Scene-biased open-set action recognizers fail in two typical
    scenarios: \myblue{known} actions in \myred{unfamiliar} scenes, and
    \myred{unknown} actions in \myblue{familiar} scenes. The former leads to low
    precision on open-set detection, while the latter leads to low recall. Our method focuses on mitigating the scene bias to improve OSAR.}
    \label{fig:motivation}
\end{figure}

Open-set action recognition (OSAR) is especially vulnerable to the spurious
information for two main reasons:
(1)~current benchmark datasets are found to be severely biased, and action
classification using 
non-action information (\eg{}, scene, object, or human) achieves high accuracy~\cite{li2018resound};
(2)~without a specific module design, the model tends to focus on static
information learning instead of temporal action
modeling~\cite{zhao2018recognize,crasto2019mars,shou2019dmc,wang2019hallucinating,piergiovanni2019representation}.

This paper focuses on mitigating the scene bias in OSAR: we speculate that current
OSAR methods are biased toward the scene, and the performance degrades when the
testing set exhibits different scene distributions from the training set.
Specifically, existing methods may fail in two typical scenarios: known action
in unfamiliar scene and unknown action in familiar scene, as illustrated
in~\cref{fig:motivation}.
For the former scenario, a scene-biased recognizer would falsely recognize the
action as unknown given the scene is unfamiliar to the training set, and lowers
the OSAR precision.
For the latter scenario, a scene-biased recognizer may falsely recognize the
unknown action as known if a familiar scene has appeared during training, which further
lowers the OSAR recall.
Consequently, the two above situations degrade the overall OSAR
performance.
To verify our speculations, a quantitative scene bias analysis experiment is
carried out in~\cref{sec:motivation}, and the results reveal a strong correlation
between the testing scene distribution shift and OSAR performance.

To mitigate scene bias, we propose a \textit{Scene-debiasing Open-set Action Recognition
method} (\abbrmethod{}), which features an \textit{adversarial scene reconstruction module}
(\reconmod{}) and an \textit{adaptive adversarial scene classification module}
(\sclsmod{}).
As shown in \cref{fig:framework}, we formulate the OSAR task as an uncertainty
estimation problem, where the recent evidential deep learning is leveraged to
quantify the second-order prediction
uncertainty~\cite{sensoy2018evidential,amini2020deep,bao2021evidential,li2022trustworthy}.
To mitigate scene bias, \reconmod{} promotes the backbone to reduce scene
information by applying adversarial learning between a decoder and the backbone.
Meanwhile, \sclsmod{} encourages the backbone to learn scene-invariant feature
by preventing a scene classifier from predicting the scene type of
input videos.

Specifically, for \reconmod{}, our intuition stems from the observation that reconstruction
autoencoders prioritize reconstructing the low-frequency part of the
input~\cite{hinton2006reducing}, which typically corresponds to the static scene
in the video domain.
Therefore, we regard the decoder that takes as input video feature and reconstructs
the video as a scene information extractor.
By applying adversarial learning between the decoder and the encoder, \reconmod{}
promotes the encoder (\ie, the feature backbone) to reduce scene information within the output feature.
Furthermore, to reduce the noise from reconstructing the foreground motion, we propose
background estimation and uncertainty-guided reconstruction to make the decoder
focus on background scene reconstruction, thus preserving motion information
during adversarial learning.

For \sclsmod{}, instead of only conducting video-level adversarial scene classification as
in~\cite{choi2019can}, we propose to adaptively apply weights on the background and foreground locations: higher weights on the action foreground and lower weights on the background scene,
where the background and foreground locations are determined by the learned spatio-temporal uncertainty map.
As a result, \sclsmod{} prioritizes debiasing on the foreground, and promotes scene-invariant action feature learning.

Extensive experiments performed on UCF101~\cite{soomro2012ucf101},
HMDB51~\cite{kuehne2011hmdb} and MiTv2~\cite{monfortmoments} demonstrate the
effectiveness of our proposed modules, and show our \abbrmethod{} achieves
state-of-the-art OSAR performance.
Besides, quantitative scene bias analysis experiments reveal that our
\abbrmethod{} achieves the lowest scene bias compared to previous arts.

To summarize, our contributions are threefold:
\begin{itemize}[leftmargin=*]
\item We design a quantitative experiment to analyze the scene bias of
current OSAR methods.
The results reveal a strong correlation between testing scene distribution shift
and OSAR performances.
Our \abbrmethod{} achieves the lowest scene bias while outperforming
state-of-the-art OSAR methods, demonstrating the effectiveness of our debias
method.
\item We propose an adversarial scene reconstruction module. By preventing a
decoder from reconstructing the video background from the extracted feature,
\reconmod{} forces the backbone to reduce scene information from the feature
while preserving motion information.
\item We propose an adaptive adversarial scene classification module,
which prevents a scene classification head from predicting the scene type of the video.
Benefiting from additional guidance from the learned uncertainty map, \sclsmod{}
promotes effective scene-invariant feature learning.
\end{itemize}

\section{Related work}
\label{sec:related-work}

\begin{figure*}[t]
    \centering
    \subfloat[Analysis on the \myblue{known} action in \myred{unfamiliar} scene
    scenario. The performances of OSAR methods degrade when closed-set
    testing samples exhibit unfamiliar scenes to the training set.
    \label{subfig:scene-bias-close}]{\includegraphics[width=0.475\linewidth]{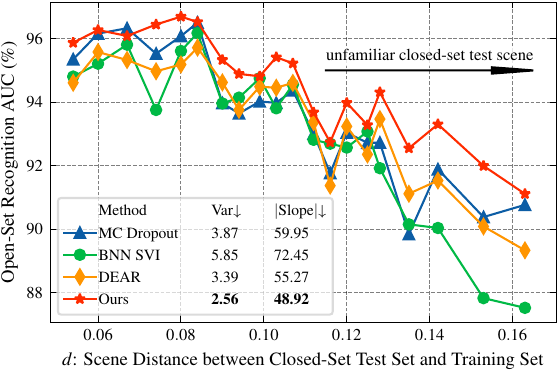}}
    \hfill
    \subfloat[Analysis on the \myred{unknown} action in \myblue{familiar} scene
    scenario. The performances of OSAR methods degrade when open-set testing
    samples exhibit familiar scenes to the training set.
    \label{subfig:scene-bias-open}]{\includegraphics[width=0.475\linewidth]{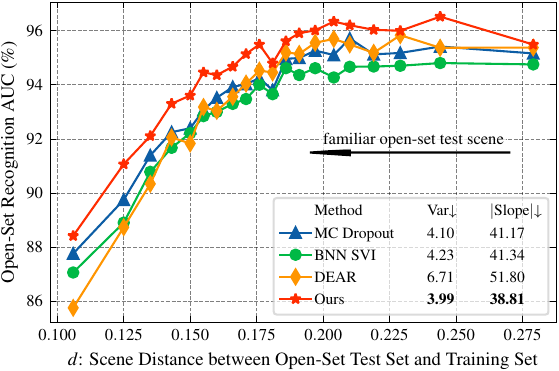}}
    \caption{Quantitative scene bias analysis using
    UCF101~\cite{soomro2012ucf101} as known and MiTv2~\cite{monfortmoments} as
    unknown. Our \abbrmethod{} is least affected by scene.}
    \label{fig:scene-bias}
\end{figure*}

\noindent\textbf{Action recognition} in the closed-set setting has been widely
exploited in recent years.
Two-stream convolutional networks~\cite{simonyan2014two} use two separate
networks to learn appearance and motion from RGB frames and optical flow,
respectively.
I3D~\cite{carreira2017quo} expands the 2D CNNs in the two-stream network to 3D
CNNs, and significantly improves recognition performance.
Due to the expensive cost of optical flow estimation, several recent
works~\cite{zhao2018recognize,crasto2019mars,shou2019dmc,wang2019hallucinating,piergiovanni2019representation}
try to learn motion information from raw videos directly.
In this paper, we also aim to learn motion information with only RGB frames
input to reduce the data process cost.

\noindent\textbf{Open-set recognition} aims to recognize testing samples
that do not belong to the training classes~\cite{scheirer2012toward}.
There are mainly two groups of work for open-set recognition, \ie{},
discriminative methods and generative methods~\cite{geng2020recent}.
For discriminative models, several traditional methods leverage support
vector machines to reject the
unknown~\cite{scheirer2014probability,jain2014multi,bendale2015towards}.
OpenMax~\cite{bendale2016towards} first adopts deep learning models in the open
set recognition problem, where it redistributes the softmax output to estimate
the uncertainty. 
DOC~\cite{shu2017doc} proposes a 1-vs-rest layer to replace the softmax layer
and tighten the decision boundary.
Recently, several methods explicitly
model the
potential open-set samples in the latent space, and promote a more
discriminative decision boundary~\cite{chen2020learning,zhou2021learning,chen2021adversarial}.
Generative methods explicitly generate samples of unknown/known classes, thus
helping learn a better decision
boundary~\cite{ge2017generative,ditria2020opengan,perera2020generative,chen2021adversarial,kong2021opengan,yue2021counterfactual,zhou2021learning,yoshihashi2019classification}.
Specifically, several
methods~\cite{yoshihashi2019classification,oza2019c2ae,sun2020conditional}
leverage the autoencoder to reconstruct the input, and use the reconstruction
error to determine open-set samples.

Most of the above methods focus on the image domain.
For the OSAR problem, ODN~\cite{shu2018odn} detects new categories by applying a
multi-class triplet thresholding method.
Busto~\etal{}~\cite{busto2018open} propose an approach for open-set domain
adaptation on action recognition.
Several
methods~\cite{krishnan2018bar,subedar2019uncertainty,krishnan2020specifying,bao2021evidential}
focus on learning the uncertainty of unknown classes.
Specifically, Bayesian neural networks are widely adopted in the action
domain~\cite{krishnan2018bar,subedar2019uncertainty,krishnan2020specifying}.
Recently, evidential deep
learning~\cite{sensoy2018evidential,amini2020deep,bao2021evidential,li2022trustworthy}
shows great potential in uncertainty estimation and achieves superior
performances~\cite{bao2021evidential} in OSAR.

\noindent\textbf{Debias} has been a challenging task in machine learning.
Previous works in the image domain include mitigating the gender
bias~\cite{bolukbasi2016man,zhao2017men,buolamwini2018gender,hendricks2018women},
and texture bias~\cite{geirhos2018imagenet}.
Several methods address this problem with adversarial
learning~\cite{xie2017controllable,zhao2018learning,elazar2018adversarial,wu2018towards,kim2019learning,choi2019can,wu2020privacy,wu2019delving},
where the label of the debias target can be extracted with off-the-shelf
pretrained models.
ContraCAM~\cite{mo2021object} alleviates the scene bias in image object, where
contrastive learning is used to automatically determine the discriminative
regions.
There also exist several debias works in the action recognition
area~\cite{li2018resound,choi2019can,bao2021evidential,hu20222tad,zhai2020two,zhai2021action,zhai2022adaptive}.
Resound~\cite{li2018resound} analyzes the scene/object/people bias existing in
current datasets.
DEAR~\cite{bao2021evidential} introduces ReBias~\cite{bahng2020learning} to
open-set action recognition, and mitigates static bias by forcing the features
learned from the original video and shuffled/static videos to be independent.
Notably, ReBias~\cite{bahng2020learning} requires simultaneously training
several backbones, while we only train one backbone with a light-weight decoder
and classification heads, greatly reducing the computational burden.
Choi~\etal{}~\cite{choi2019can} also leverages an adversarial scene
classification module; however, they conduct adversarial learning on the whole
frame instead of considering specific scene locations.
We propose a guide loss to direct the adversarial classification on the
foreground, thus to promote effective scene-invariant feature learning.

\section{The effect of the scene in OSAR}
\label{sec:motivation}

As illustrated in~\cref{fig:motivation}, we speculate that existing OSAR methods are
vulnerable to scene bias under two typical settings:
known action in unfamiliar scene and unknown action in familiar scene.
To measure the severity of existing methods affected by the two problems, we
conduct the following quantitative experiments.

\begin{figure*}[t!]
    \includegraphics[width=\linewidth]{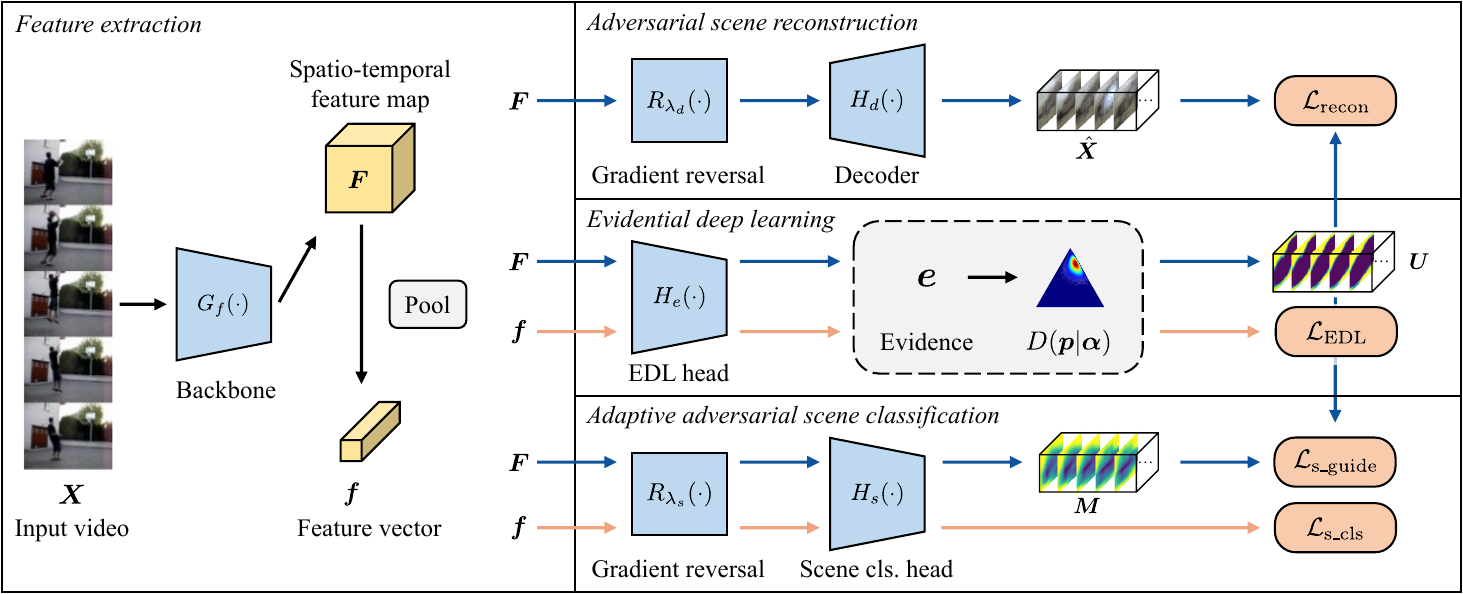}
    \caption{Framework overview. Our \abbrmethod{} consists of four major
    modules: the feature extraction module extracts spatio-temporal features
    from the input video; the evidential deep learning module estimates the
    prediction uncertainty and outputs the spatio-temporal uncertainty map; the
    adversarial scene reconstruction module (\reconmod{}) reconstructs the video
    background; the adaptive adversarial scene classification module
    (\sclsmod{}) predicts the scene in the video. The latter two modules are
    trained in an adversarial way to learn scene-invariant features.}
    \label{fig:framework}
\end{figure*}

\noindent\textbf{Settings.}
The following describes the experimental setup to analyze the first known action
in unfamiliar scene problem, and the setup for the second problem can be
conducted similarly.
Our essential goal is to build different combinations of testing sets, such that
each testing set contains different closed-set testing samples that exhibit
different scene similarities to the training set while keeping the open-set
testing samples the same, and observe how these different combinations of
testing sets affect performances of existing OSAR methods.
Specifically, we first use an off-the-shelf scene classifier to extract the
scene features $\bm{f}_\text{scene}$ on each training and testing video.
After that, for each closed-set testing video, we compute its scene feature
cosine distance to all training videos, and use the minimal distance to indicate
the scene distance between this testing video and the training set.
Subsequently, we sort all closed-set testing videos with their scene distance to
the training set, and divide them into several non-overlapping equal-sized
subsets.
For each closed-set testing subset with size $L$, we define its scene distance
to the training set $d$ as the average of minimal scene feature cosine distance
between each testing video and all training videos:
\begin{equation}
    d = \frac{1}{L} \sum_{i=1}^L \min_j \left (1 - \bm{u}_i\bm{v}_j \right ),
\end{equation}
where the unit vector $\bm{u}_i$, and $\bm{v}_j$ are normalized scene feature $\bm{f}_\text{scene}$ of the $i$-th testing video and the $j$-th training video, respectively.
Finally, we fix the open-set testing set and combine it with different
closed-set testing subsets, and observe how the performances change.
Note that we additionally ensure that each closed-set subset is class-balanced, such
that all testing set combinations achieve the same
openness~\cite{scheirer2012toward}, which measures how open the testing
environment is.

\noindent\textbf{Datasets and evaluation.}
We perform the experiments with the UCF101 training set for
training~\cite{soomro2012ucf101}, UCF101 validation set as closed-set testing
set and MiTv2 validation set as open-set testing set~\cite{monfortmoments}.
Two metrics are used to quantify the scene bias: the variance of the OSAR AUCs
under different testing combinations, and the absolute value of the linear
fitting slope of the performance change curves.
We note that all methods exhibit Pearson correlation coefficients between OSAR
AUC and $d$ larger than $0.7$, which indicates a strong linear correlation and
justifies the use of linear slope for evaluation.
We divide the closed-set/open-set testing set into $20$ subsets to conduct the
two evaluations, respectively.

\noindent\textbf{Analysis.}
Our \abbrmethod{} is compared to previous
methods~\cite{gal2016dropout,krishnan2018bar,bao2021evidential}
in \cref{fig:scene-bias}.
The analysis of known action in unfamiliar scene is presented in
\cref{subfig:scene-bias-close}.
A clear performance decrease trend is observed as the scene distance between the
closed-set testing set and the training set increases.
The results suggest that current OSAR methods rely on the scenes to make
predictions: known actions with familiar scenes are easier to recognize, while
those with unfamiliar scenes are harder to recognize.
\cref{subfig:scene-bias-open} analyzes the unknown action in familiar scene
counterpart, where we also observe a clear increasing trend as the scene
distance between the open-set testing set and the training set increases,
indicating unknown actions with unfamiliar scenes are easier to recognize.
Both figures show a strong correlation between the scene distance and the OSAR
performance, suggesting that the scene is an essential cue for open-set recognition.
Moreover, we find that our \abbrmethod{} achieves the lowest variance and absolute
slope, showing its scene-debiasing capability.

\section{Method}
\label{sec:method}

The overview of our proposed \abbrmethod{} is illustrated
in~\cref{fig:framework}.
Given an input video, \abbrmethod{} predicts an uncertainty score that
measures how likely this video contains known actions that are used for
training.
To mitigate scene bias, we aim to suppress the performance of scene-related
tasks (reconstruction and classification) while maintaining the action recognition performance.
This reduces the scene information in the learned debiased representation, such that the following uncertainty
estimation process will be less dependent on the scene.

\subsection{Evidential deep learning}
To distinguish the known and unknown samples, a scoring function is needed to
measure the likelihood that the samples are unknown.
To this end, we leverage the recent evidential deep learning (EDL)
methods~\cite{dempster1968generalization,josang2016subjective,sensoy2018evidential,amini2020deep,bao2021evidential,li2022trustworthy}
for uncertainty estimation, which mitigates the
over-confident~\cite{moon2020confidence,van2020uncertainty} and computationally
costly~\cite{blundell2015weight,gal2016dropout,corbiere2019addressing} problems
of existing uncertainty estimators.
Essentially, for the $C$-way classification, EDL first collects the evidence
that supports the given sample to be classified into a particular class and then
builds a Dirichlet class probability distribution parameterized over the
evidence.
The resulting distribution models the second-order class probabilities and
uncertainty.
We refer readers unfamiliar with EDL to~\cite{sensoy2018evidential}.

Specifically, denote the input video as $\bm{X} \in \mathbb{R}^{H\times W\times
T\times D}$, where $H, W, T$ and $D$ represent height, width, number of frames
and channels, respectively.
The backbone $G_f(\cdot)$ maps it into a spatio-temporal feature map
$\bm{F} \in \mathbb{R}^{H'\times W'\times T'\times D'}$, which is further
average pooled as a feature vector $\bm{f} \in \mathbb{R}^{D'}$.
The EDL head $H_e(\cdot)$ takes as input the feature vector
$\bm{f}$, and predicts a non-negative evidence vector $\bm{e} = H_e(\bm{f}) \in
\mathbb{R}^{C}_{\geq 0}$, which parameterizes the following Dirichlet class
probability distribution:
\begin{equation}
    D(\bm{p}|\bm{\alpha}) = 
    \left \{
        \begin{array}{ll}
            \frac{1}{B(\bm{\alpha})} \Pi_{j=1}^C p_j^{\alpha_j - 1} &
            \text{for}\ \bm{p}\in \mathcal{S}_C, \\
            0 & \text{otherwise},
        \end{array}
    \right.
\end{equation}
where $\mathcal{S}_C$ is the $C$-dim unit simplex, $\alpha_j = e_j + 1$, and
$B(\bm{\alpha})$ is the $C$-dim multinomial beta
function~\cite{sensoy2018evidential}.
Thus, the cross entropy action classification loss reduces to the following:
\begin{equation}
    \mathcal{L}_{\mathrm{EDL}} = \sum_{i=1}^{C} y_{i} \big( \log S - \log
    \alpha_{i} \big),
\end{equation}
where $\bm{y}$ is the one-hot label vector, $S=\sum_{i=1}^C \alpha_i$ is the
the total strength of the Dirichlet distribution.
During inference, the class probability is given as the mean of the Dirichlet
distribution $\bm{p} = \bm{\alpha}/S$, and the prediction uncertainty is
deterministically given as $u=C/S$.
As the EDL head estimates the uncertainty relying on the feature vector
$\bm{f}$, we aim to reduce scene information in $\bm{f}$, so that the uncertainty
estimation process is less dependent on the scene.
Scene debias is accomplished by the following two modules via adversarial training.

\subsection{Adversarial scene reconstruction}
We take inspiration from reconstruction-based video anomaly detection methods,
where locations with abnormal motions typically incur high reconstruction
errors~\cite{ramachandra2020survey}.
In action recognition, we empirically find that such reconstruction
prioritizes reconstructing static background scene, while achieving low reconstruction
quality on the action-related foreground.
Thus, our \reconmod{} adds a decoder to reconstruct the video background.
By regarding the decoder as a scene information extractor, we force the learned feature
$\bm{F}$ not to contain scene information to hinder scene reconstruction in an
adversarial learning manner.

Specifically, given spatio-temporal feature tensor $\bm{F}$, we feed it into a
decoder $H_d(\cdot)$ to reconstruct the raw video frames: $\hat{\bm{X}} =
H_d(R_{\lambda_d}(\bm{F}))$.
$R_{\lambda_d} (\cdot)$ is a gradient reversal
layer~\cite{ganin2015unsupervised,choi2019can} that acts as an identity function
during forward propagation, and reverses the gradient by a factor of $\lambda_d$
during backward propagation: $\frac{\mathrm{d}R_{\lambda_d}
(\bm{X})}{\mathrm{d}\bm{X}} = -\lambda_d \bm{I}$, where $\bm{I}$ is an identity
matrix.
In this way, the reconstruction loss is adversarial in that it forces the
backbone $G_f(\cdot)$ to reduce the scene information contained in the output
feature $\bm{F}$ to hinder reconstruction (\ie{}, maximize the loss), while
encouraging the decoder to extract the scene information from the feature for
reconstruction (\ie{}, minimize the loss).
Despite its simplicity, such adversarial reconstruction inevitably loses motion
information, as the action-related foreground is also involved in this process.
We address this problem by enforcing the decoder to focus on reconstructing the
background with two additional designs: background estimation and 
uncertainty-weighted reconstruction.

\noindent\textbf{Background estimation.}
Instead of using raw frames $\bm{X}$ as the reconstruction target, we
propose to use the video background $\bar{\bm{X}}$ for the adversarial
reconstruction, such that the foreground action information will not be removed
from the feature.
To achieve background estimation, we leverage the temporal median filter (TMF),
which has been demonstrated as an effective background estimation
method~\cite{mandal2020motionrec,tezcan2020bsuv,sahoo2021contrast}.
Specifically, for a given pixel location, the most frequently repeated intensity
in a sequence of frames is most likely to be the background value for that
scene ~\cite{piccardi2004background,liu2016scene}.
Thus, TMF takes the pixel-wise temporal median in a sliding window on a frame
sequence as the corresponding background.
We denote the background clip of the video as $\bar{\bm{X}}$, which is used as
the reconstruction target.

\noindent\textbf{Uncertainty-weighted reconstruction.}
Despite the simplicity and effectiveness of TMF, it extracts inferior background
in videos with static foreground (\eg{}, apply eye makeup), which further disturbs adversarial scene reconstruction.
To address this problem, we leverage the spatio-temporal uncertainty map.
Similarly to the class activation map~\cite{zhou2016learning}, we apply the EDL
head onto the spatio-temporal feature $\bm{F}$ to generate a spatio-temporal
evidence map $\bm{E} = H_e({\bm{F}}) \in \mathbb{R}^{H'\times W'\times T'\times
C}$.
The evidence map $\bm{E}$ can be converted to a spatio-temporal uncertainty map
$\bm{U} \in \mathbb{R}^{H'\times W'\times T'}$ according to
DST~\cite{dempster1968generalization}: $u_{i, j, t} = C / \sum_{c} (e_{i, j, t,
c} + 1)$, where $u_{i, j, t}$ is the element of $\bm{U}$ at index $i, j, t$.

Intuitively, similar to the class activation map that indicates discriminative
locations that respond to the class
label~\cite{pathak2014fully,pinheiro2015image}, 
the obtained uncertainty map is expected to indicate locations that are discriminative for action recognition (\ie{}, foreground) with low uncertainty, while high uncertainty indicates the background scene.
Meanwhile, considering that the reconstruction task should focus on the background scene
while neglecting the foreground action, the uncertainty map can serve as a weight map to
guide the reconstruction.
Specifically, scene locations with high uncertainties are assigned higher weights
for reconstruction, so that the backbone $G_f(\cdot)$ will focus on removing
information at these locations to disturb reconstruction.
The final reconstruction loss $\mathcal{L}_{\text{recon}}$ is formulated as a
weighted L1 loss:
\begin{equation}
    \mathcal{L}_{\text{recon}} = \frac{1}{HWTD} \sum_{i, j, t, d} u'_{i,j,t}\|
    \bar{x}_{i,j,t,d} - \hat{x}_{i,j,t,d}\|_1,
    \label{eq:l_recon}
\end{equation}
where $u'_{i, j, t}$ is the element of $\bm{U}'$ at index $i, j, t$ defined in
\cref{eg:u'}.
Specifically, since $\bm{U}$ has different spatio-temporal resolution from our
reconstruction target $\bar{\bm{X}}$, we upsample it to have the same size as
$\bar{\bm{X}}$.
Additionally, min-max normalization is applied on $\bm{U}$, such that $\bm{U}'$
ranges from $0$ to $1$, meaning that the most confident locations will have no
reconstruction loss, while the most uncertain locations have the largest
reconstruction weight as $1$.
These steps are formulated as follows:
\begin{equation}
    \bm{U}' = \text{up} \left( \text{norm} (\bm{U}) \right),
    \label{eg:u'}
\end{equation}
where $\text{up}(\cdot)$ is the trilinear interpolation upsampling function, and
$\text{norm}(\cdot)$ is the min-max normalization.

\subsection{Adaptive adversarial scene classification}
To further facilitate scene-invariant action feature learning, we propose
\sclsmod{} for adaptive adversarial video scene classification.
Denote the scene label as $\bm{y}_\text{s} \in \mathbb{R}^{N}$, where $N$ is the
number of pre-defined scene classes.
The scene classification head $H_s (\cdot)$ predicts the video-level scene type
given feature vector $\bm{f}$ as $\hat{\bm{y}}_{\text{s}} =
H_s(R_{\lambda_{s}}(\bm{f}))$, where $\lambda_s$ is the gradient reversal
weight.
The adversarial scene classification is achieved via a cross-entropy
classification loss $\mathcal{L}_{\text{s\_cls}}$:
\begin{equation}
    \mathcal{L}_{\text{s\_cls}} = -\frac{1}{N} \sum_{i=1}^N y_{\text{s}, i} \log
    \frac{\exp(\hat{y}_{\text{s}, i})} {\sum_{j=1}^N \exp(\hat{y}_{\text{s},
    j})}.
\end{equation}

Despite previous exploration~\cite{choi2019can}, we note that blindly performing adversarial scene classification on the whole video may yield suboptimal
OSAR results.
As video scene classification tends to focus on static
cues~\cite{li2018resound}, this can cause the action foreground to be
disregarded during the adversarial classification, hindering the learning of
scene-invariant action feature.
This issue becomes more pronounced when there is a strong correlation between
the action foreground and the scene.
To mitigate this problem, we propose to direct the adversarial scene
classification towards the foreground locations.

In our \sclsmod{}, we use the uncertainty map $\bm{U}$ to adaptively guide the
learning of scene classification, so that the adversarial classification
focuses on the foreground locations.
Specifically, the scene class activation map $\bm{M} \in \mathbb{R}^{H'\times
W'\times T'}$ can be obtained by passing the feature map $\bm{F}$ to the scene
classification head $m_{i,j,t} = H_s(R_{\lambda_s} (\bm{F}))_{i,j,t,n}$, where
$n = \mathop{\mathrm{argmax}}\limits_{i} y_{\text{s},i}$.
The uncertainty guidance is accomplished by maximizing the difference between the
normalized uncertainty map and the scene class activation map.
As both terms are within range $[0, 1]$, we minimize the L1 distance between
$1-\text{norm}(\bm{U})$ and $\text{norm}(\bm{M})$ as a proxy:
\begin{equation}
    \mathcal{L}_{\text{s\_guide}} = \frac{1}{HWT} \sum_{i,j,t} \| 1 - \text{norm}
    (\bm{U})_{i,j,t} - \text{norm} (\bm{M})_{i,j,t} \|_1.
\end{equation}

\subsection{Model training}

The overall training loss $\mathcal{L}$ is a weighted sum of the evidential
learning loss $\mathcal{L}_{\text{EDL}}$, adversarial scene reconstruction loss
$\mathcal{L}_{\text{recon}}$, adversarial scene classification loss
$\mathcal{L}_{\text{s\_cls}}$ and the scene guide loss
$\mathcal{L}_{\text{s\_guide}}$:
\begin{equation}
    \mathcal{L} = \mathcal{L}_{\text{EDL}} + w_{\text{recon}}
    \mathcal{L}_{\text{recon}} + w_{\text{s\_cls}} \mathcal{L}_{\text{s\_cls}} +
    w_{\text{s\_guide}} \mathcal{L}_{\text{s\_guide}} 
\end{equation}
where $w_{\text{recon}}$, $w_{\text{s\_cls}}$ and $w_{\text{s\_guide}}$ are
weight hyperparameters.

\begin{table*}[t!]
    \centering
    \resizebox{\linewidth}{!}{%
    \begin{tabular}{c|c|c|c|c|c|c|c|c|c}
        \hline 
        \multirow{2}{*}{Methods} & \multicolumn{4}{c|}{UCF101~\cite{soomro2012ucf101}+MiTv2~\cite{monfortmoments}} & \multicolumn{4}{c|}{UCF101~\cite{soomro2012ucf101}+HMDB51~\cite{kuehne2011hmdb}} & Closed-set\tabularnewline
        \cline{2-9} \cline{3-9} \cline{4-9} \cline{5-9} \cline{6-9} \cline{7-9} \cline{8-9} \cline{9-9} 
         & AUC $\uparrow$ & FAR@95 $\downarrow$ & TPR@10 $\uparrow$ & Open maF1 $\uparrow$ & AUC $\uparrow$ & FAR@95 $\downarrow$ & TPR@10 $\uparrow$ & Open maF1 $\uparrow$ & Accuracy\tabularnewline
        \hline 
        SoftMax & 44.47 & 96.93 & 8.85 & $55.50\pm0.45$ & 44.34 & 97.91 & 3.66 & $73.13\pm0.12$ & 94.10\tabularnewline
        OpenMax~\cite{bendale2016towards} & 63.96 & 45.89 & 3.78 & $66.21\pm0.16$ & 63.67 & 80.53 & 6.54 & $67.81\pm0.12$ & 56.54\tabularnewline
        MC Dropout~\cite{gal2016dropout} & 93.66 & 25.43 & 85.72 & $68.12\pm0.20$ & 86.11 & 77.50 & 70.13 & $71.13\pm0.15$ & 94.13\tabularnewline
        BNN SVI~\cite{krishnan2018bar} & 93.16 & 25.88 & 79.36 & $67.96\pm0.19$ & 85.63 & 71.52 & 66.14 & $71.57\pm0.17$ & 93.89\tabularnewline
        DEAR~\cite{bao2021evidential} & 93.52 & 29.53 & 84.03 & $75.12\pm0.27$ & 87.12 & 71.32 & 72.21 & $88.07\pm0.20$ & 93.97\tabularnewline
        \hline 
        \abbrmethod{} (Ours) & \textbf{94.60} & \textbf{25.33} & \textbf{86.47} & $\mathbf{76.22\pm0.32}$ & \textbf{88.10} & \textbf{69.57} & \textbf{72.75} & $\mathbf{89.55\pm0.22}$ & \textbf{95.24}\tabularnewline
        \hline 
    \end{tabular}%
    }
    \caption{Comparison with state-of-the-art methods. All methods are trained
    on UCF101~\cite{soomro2012ucf101}, and evaluated on two different open sets
    where unknown samples are from HMDB51~\cite{kuehne2011hmdb} and
    MiTv2~\cite{monfortmoments}, respectively. Performances with different backbones are listed in the appendix.}
    \label{tab:comparison-with-sota}
\end{table*}

\section{Experiments}
\label{sec:experiments}

\noindent\textbf{Datasets}.
We follow DEAR~\cite{bao2021evidential} to use three datasets for evaluation:
UCF101~\cite{soomro2012ucf101}, HMDB51~\cite{kuehne2011hmdb} and
MiTv2~\cite{monfortmoments}.
We use the training split 1 from UCF101 for training, which consists of $9,537$
videos from $101$ classes.
For testing, validation split one from UCF101 is used as known samples, and
testing split 1 of HMDB51 and the testing set of MiTv2 are respectively used as
unknown samples.
The HMDB51 testing set consists of $1,530$ videos from $51$ classes, thus the
openness of testing combination UCF101 + HMDB51 is $10.6\%$; the MiTv2 testing
set consists of $30,500$ videos from $305$ classes, achieving an openness of
$36.9\%$.
Note that openness measures how open the testing environment is, and increases as the
number of testing classes increases~\cite{scheirer2012toward}.
We only use the MiTv2 dataset for the scene bias evaluation, as reported in
\cref{sec:motivation}.
Note that HMDB51 is a relatively small dataset, which prevents it from splitting
into multiple subsets to perform the scene bias evaluation.

\noindent\textbf{Evaluation metrics}.
We use accuracy on the closed testing set for closed-set classification
evaluation.
For binary open-set recognition, we use the area under the receiver operating
characteristic curve (AUC), false alarm rate at a true positive rate of $95\%$
(FAR@95) and the true positive rate at a false positive rate of $10\%$ (TPR@10)
for evaluation.
For the C + 1 way classification (\ie{}, the C known classes and the unknown
class), we follow DEAR~\cite{bao2021evidential} to report the mean and variance
of open macro F1 (open maF1), which weighted sums the macro F1 for the C+1 way
classification under different openness points.
Note that open macro F1 is a threshold-dependent metric and the uncertainty threshold
is set to the maximal training uncertainty.
We use AUC as the main metric, as FAR@95 and TPR@10 are only applicable for
particular points on the ROC curve, while open maF1 is sensitive to the
threshold value.

\begin{figure}[t!]
    \subfloat[SoftMax]{\includegraphics[width=0.475\linewidth]{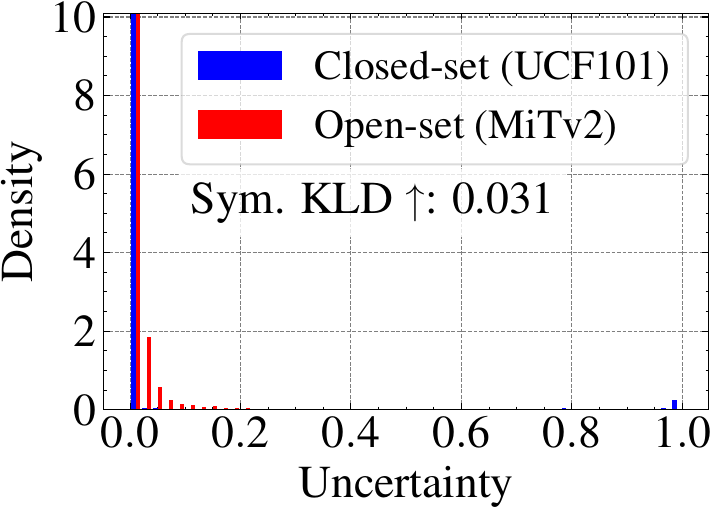}}
    \hfil
    \subfloat[OpenMax~\cite{bendale2016towards}]{\includegraphics[width=0.475\linewidth]{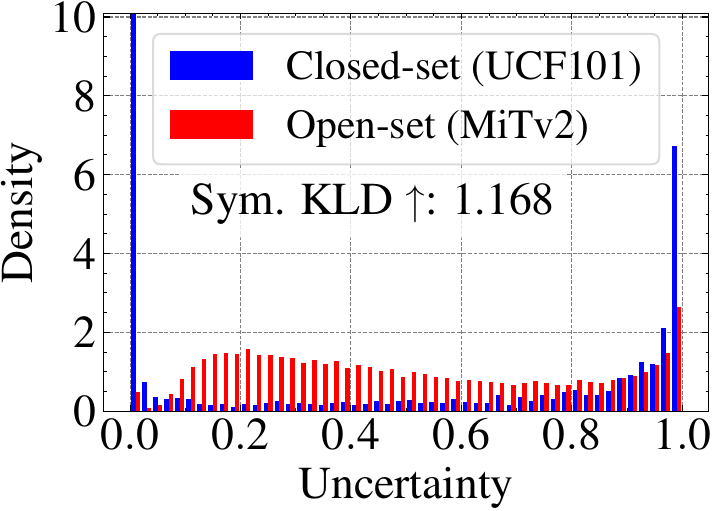}}
    
    \subfloat[MC Dropout~\cite{gal2016dropout}]{\includegraphics[width=0.475\linewidth]{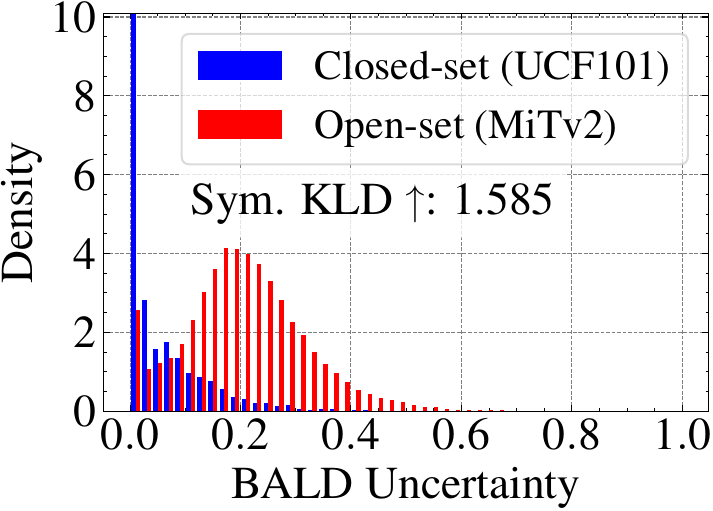}}
    \hfil
    \subfloat[BNN SVI~\cite{krishnan2018bar}]{\includegraphics[width=0.475\linewidth]{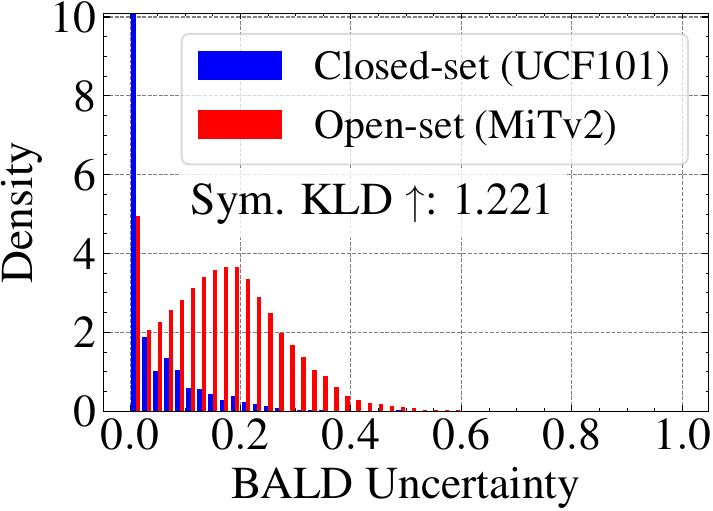}}
    
    \subfloat[DEAR~\cite{bao2021evidential}]{\includegraphics[width=0.475\linewidth]{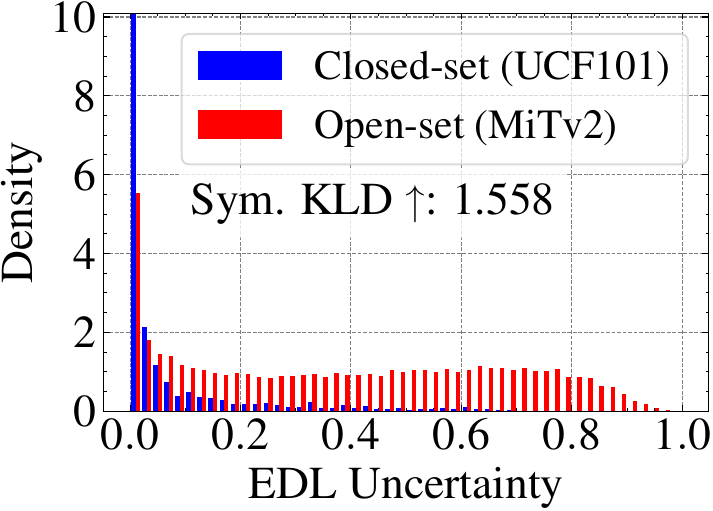}}
    \hfil
    \subfloat[\abbrmethod{} (Ours)]{\includegraphics[width=0.475\linewidth]{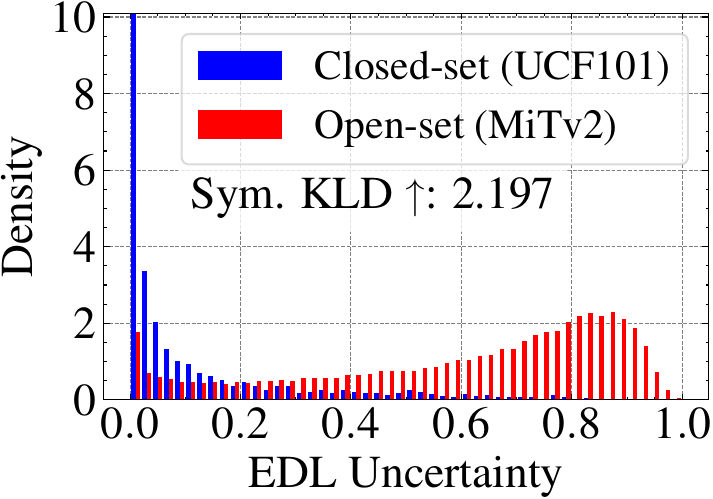}}
    \caption{Uncertainty distributions visualization on
    UCF101~\cite{soomro2012ucf101} + MiTv2~\cite{monfortmoments}. Our method
    achieves the best open-set and closed-set uncertainty separation with the
    highest symmetric KLD. Uncertainties are normalized to $[0,1]$ for better
    visualization.}
    \label{fig:uncertainty-distribution}
\end{figure}

\begin{table*}[t]
    \centering
    \resizebox{\linewidth}{!}{%
    \begin{tabular}{cc|c|c|c|c|c|c|c|c|c|c|c|c}
        \hline 
        \multirow{2}{*}{\reconmod{}} & \multirow{2}{*}{\sclsmod{}} & \multicolumn{4}{c|}{UCF101~\cite{soomro2012ucf101}+MiTv2~\cite{monfortmoments}} & \multicolumn{4}{c|}{UCF101~\cite{soomro2012ucf101}+HMDB51~\cite{kuehne2011hmdb}} & \multicolumn{2}{c|}{KAUS} & \multicolumn{2}{c}{UAFS}\tabularnewline
        \cline{3-14} \cline{4-14} \cline{5-14} \cline{6-14} \cline{7-14} \cline{8-14} \cline{9-14} \cline{10-14} \cline{11-14} \cline{12-14} \cline{13-14} \cline{14-14} 
         &  & AUC $\uparrow$ & FAR@95 $\downarrow$ & TPR@10 $\uparrow$ & Open maF1 $\uparrow$ & AUC $\uparrow$ & FAR@95 $\downarrow$ & TPR@10 $\uparrow$ & Open maF1 $\uparrow$ & Var $\downarrow$ & $|$Slope$|$ $\downarrow$ & Var $\downarrow$ & $|$Slope$|$ $\downarrow$\tabularnewline
        \hline 
        - & - & 91.73 & 28.84 & 78.96 & $68.55\pm0.34$ & 85.63 & 78.59 & 68.10 & $87.73\pm0.22$ & 6.12 & 75.51 & 6.17 & 75.52\tabularnewline
        \checkmark & - & 94.13 & 27.52 & 85.72 & $73.49\pm0.35$ & 87.49 & \textbf{69.41} & 72.31 & $89.52\pm0.21$ & 3.82 & 59.20 & 4.16 & 49.20\tabularnewline
        - & \checkmark & 93.58 & 26.43 & 83.16 & $72.16\pm0.30$ & 87.22 & 71.45 & 69.80 & $87.47\pm0.19$ & 4.43 & 63.62 & 4.49 & 63.63\tabularnewline
        \checkmark & \checkmark & \textbf{94.60} & \textbf{25.33} & \textbf{86.47} & $\mathbf{76.22\pm0.32}$ & \textbf{88.10} & 69.57 & \textbf{72.75} & $\mathbf{89.55\pm0.22}$ & \textbf{2.56} & \textbf{48.92} & \textbf{3.99} & \textbf{38.81}\tabularnewline
        \hline 
    \end{tabular}%
    }
    \caption{Ablation study on the proposed \reconmod{} and \sclsmod{}. The last
    four columns analyzes the scene bias under the known action in unfamiliar
    scene (KAUS) and the unknown action in familiar scene (UAFS) scenarios.}
    \label{tab:main-results}
\end{table*}

\noindent\textbf{Implementation details.}
Our method is implemented with MMAction2~\cite{2020mmaction2}, a toolbox based
on PyTorch~\cite{pytorch}.
Kinetics400~\cite{carreira2017quo} pre-trained ResNet50-based
I3D~\cite{he2016deep,carreira2017quo} is adopted as the backbone.
We follow~\cite{choi2019can} to use off-the-shelf
Places365~\cite{zhou2017places} pretrained ResNet50~\cite{he2016deep} to extract
video scene feature and label.
We implement the evidential learning head as a single-layer MLP followed by ReLU
activation following DEAR~\cite{bao2021evidential}, implement the decoder as
five consecutive 3D transpose convolutional layers and implement the scene
recognition head as a five-layer MLP following~\cite{choi2019can}.
We use the SGD optimizer with an initial learning rate of $0.001$, which
decreases by a factor of $0.1$ for every $20$ epochs with a total epoch of $50$.
All hyperparameters are determined via a grid search: $\lambda_d = 1$,
$\lambda_s = 10$, $w_{\text{recon}} = w_{\text{s\_cls}} = 1$, and
$w_{\text{s\_guide}} = 0.1$.

\subsection{Comparison with the state-of-the-art}

Our \abbrmethod{} is superior to previous methods in two aspects: lower scene
bias and higher performance.

\noindent\textbf{Scene bias evaluation} is analyzed in \cref{fig:scene-bias},
where two typical scenarios are evaluated: known action in unfamiliar scene and
unknown action in familiar scene.
Our \abbrmethod{} achieves the lowest variance and absolute slope in both
scenarios, showing that our method is least affected by the scene bias compared to
previous methods.
Notably, the performance improvement is more significant when the testing scene
distribution is distinct from the training (\ie{}, right part of
\cref{subfig:scene-bias-close} and left part of \cref{subfig:scene-bias-open}).
\emph{Similar trends are also observed when using Open maF1 for the scene bias
evaluation as well as using different backbones (details are in the appendix).}
Furthermore, we show that our method surpasses several debias
methods~\cite{choi2019can,bahng2020learning,mo2021object} in the appendix.
Such results demonstrate that our \abbrmethod{} learns better scene-invariant
action features and strong scene-debiasing capability.

\noindent\textbf{Performance comparison with the state-of-the-art} is listed in
\cref{tab:comparison-with-sota}, where both OSAR and closed-set classification
performances are reported.
The results reveal that our \abbrmethod{} outperforms all previous methods under all
metrics in both OSAR and closed-set classification tasks.
Furthermore, we visualize the uncertainty distributions in
\cref{fig:uncertainty-distribution}, where the separation between closed-set and
open-set uncertainties is quantified with symmetric Kullback-Leibler divergence
(sym. KLD).
We observe that our \abbrmethod{} generates a noticeable bimodal distribution and the
highest sym.~KLD between closed-set and open-set uncertainties.
\emph{We further show our \abbrmethod{} achieves state-of-the-art OSAR
performance with different backbones in the appendix.}
Such a clear performance advantage demonstrates the effectiveness of our method.

\subsection{Ablation studies}

\cref{tab:main-results} summarizes the ablation studies on \reconmod{} and
\sclsmod{}.
We have the following two observations.
(1) Both modules individually improve the performance over the EDL baseline, and
the combination of them leads to better performance, validating their effectiveness individually and complementarily.
\emph{Notably, with only \reconmod{}, our method outperforms all previous OSAR
methods in terms of AUC.}
(2) The performance improvement from \sclsmod{} is lower than that from
\reconmod{}.
We speculate that this is because the predicted scene label may be noisy and mislead adversarial learning.
(3) The last four columns of \cref{tab:main-results} list the scene bias
analysis, showing that both modules alleviate the scene bias.

\begin{table}[t]
    \centering
    \resizebox{0.85\linewidth}{!}{%
    \begin{tabular}{c|cc|cc}
        \hline 
        \multirow{2}{*}{Method} & \multicolumn{2}{c|}{Biased (Kinetics~\cite{carreira2017quo})} & \multicolumn{2}{c}{Unbiased (Mimetics~\cite{weinzaepfel2021mimetics})}\tabularnewline
         & Top1 $\uparrow$ & Top5 $\uparrow$& Top1 $\uparrow$& Top5 $\uparrow$\tabularnewline
        \hline 
        EDL Baseline & 91.11 & 99.27 & 25.32 & 69.62\tabularnewline
        DEAR~\cite{bao2021evidential} & 91.18 & 99.54 & 34.58 & 75.00\tabularnewline
        \abbrmethod{} (Ours) & \textbf{92.37} & \textbf{99.69} & \textbf{37.18} & \textbf{78.92}\tabularnewline
        \hline 
    \end{tabular}%
    }
    \caption{Classification accuracy on biased and unbiased datasets.}
    \label{tab:ablation-kinetics-mimetics}
\end{table}

\begin{table}[t]
    \centering
    \resizebox{0.85\linewidth}{!}{%
    \begin{tabular}{cc|c|c|c}
        \hline 
        \multirow{2}{*}{\reconmod{}} & \multirow{2}{*}{\sclsmod{}} & \multicolumn{3}{c}{CKA~\cite{kornblith2019similarity} $\downarrow$}\tabularnewline
        \cline{3-5} \cline{4-5} \cline{5-5} 
         &  & UCF101~\cite{soomro2012ucf101} & HMDB51~\cite{kuehne2011hmdb} & MiTv2~\cite{monfortmoments}\tabularnewline
        \hline 
        - & - & 0.34 & 0.41 & 0.37\tabularnewline
        \checkmark & - & 0.27 & 0.37 & 0.34\tabularnewline
        - & \checkmark & \textbf{0.23} & 0.30 & 0.28\tabularnewline
        \checkmark & \checkmark & \textbf{0.23} & \textbf{0.28} & \textbf{0.27}\tabularnewline
        \hline 
    \end{tabular}%
    }
    \caption{Feature similarity between the learned action feature $\bm{f}$ and
    the scene feature $\bm{f}_{\text{scene}}$ on the closed-set testing set
    (UCF101~\cite{soomro2012ucf101}) and the open-set testing sets
    (HMDB51~\cite{kuehne2011hmdb} and MiTv2~\cite{monfortmoments}). The
    similarity is measured with centered kernel alignment
    (CKA)~\cite{kornblith2019similarity}.}
    \label{tab:ablation-feature-similarity}
\end{table}

\begin{table}[t]
    \centering
    \resizebox{\linewidth}{!}{%
    \begin{tabular}{ccc|c|c|c|c}
        \hline 
        \multirow{2}{*}{\reconmod{}} & Bg. & Unc. & \multirow{2}{*}{AUC $\uparrow$} & \multirow{2}{*}{FAR@95 $\downarrow$} & \multirow{2}{*}{TPR@10 $\uparrow$} & \multirow{2}{*}{Open maF1 $\uparrow$}\tabularnewline
         & Est. & Weight &  &  &  & \tabularnewline
        \hline 
        - & - & - & 91.73 & 28.84 & 78.96 & $68.55\pm0.34$\tabularnewline
        \checkmark & - & - & 92.12 & 28.67 & 79.69 & $69.38\pm0.34$\tabularnewline
        \checkmark & - & \checkmark & 93.66 & 27.59 & 82.13 & $72.46\pm0.32$\tabularnewline
        \checkmark & \checkmark & - & 92.73 & 28.33 & 81.84 & $71.58\pm0.33$\tabularnewline
        \checkmark & \checkmark & \checkmark & \textbf{94.13} & \textbf{27.52} & \textbf{85.72} & $\mathbf{73.49\pm0.35}$\tabularnewline
        \hline 
    \end{tabular}%
    }
    \caption{Ablation study on the adversarial reconstruction on
    UCF101~\cite{soomro2012ucf101} + MiTv2~\cite{monfortmoments} datasets.}
    \label{tab:abaltion-on-adversarial-reconstruction}
\end{table}

\noindent\textbf{Representation debiasing} is analyzed in two aspects:
out-of-distribution (OOD) generalization ability and feature similarity between
the learned action feature and the scene feature.
The OOD generalization is compared in~\cref{tab:ablation-kinetics-mimetics},
where we follow DEAR~\cite{bao2021evidential} to use $10$ classes on Kinetics
for training and biased testing, and the same categories from
Mimetics~\cite{weinzaepfel2021mimetics} for unbiased testing.
The results reveal that our \abbrmethod{} outperforms the EDL baseline and
DEAR~\cite{bao2021evidential} in both settings, showing our stronger debias
capability.
We further compare the feature similarity between the learned action feature and
the scene feature in~\cref{tab:ablation-feature-similarity} with centered kernel
alignment (CKA)~\cite{kornblith2019similarity}, which measures the learned
representation similarity between models trained on different datasets.
CKA is in range $[0, 1]$, and larger value indicates higher similarity.
The results reveal our modules successfully reduce the similarity between the
learned action feature $\bm{f}$ and the video scene feature
$\bm{f}_{\text{scene}}$ on all testing sets, showing our method reduces the
scene information in the extracted feature.

\begin{figure}[t!]
    \includegraphics[width=\linewidth]{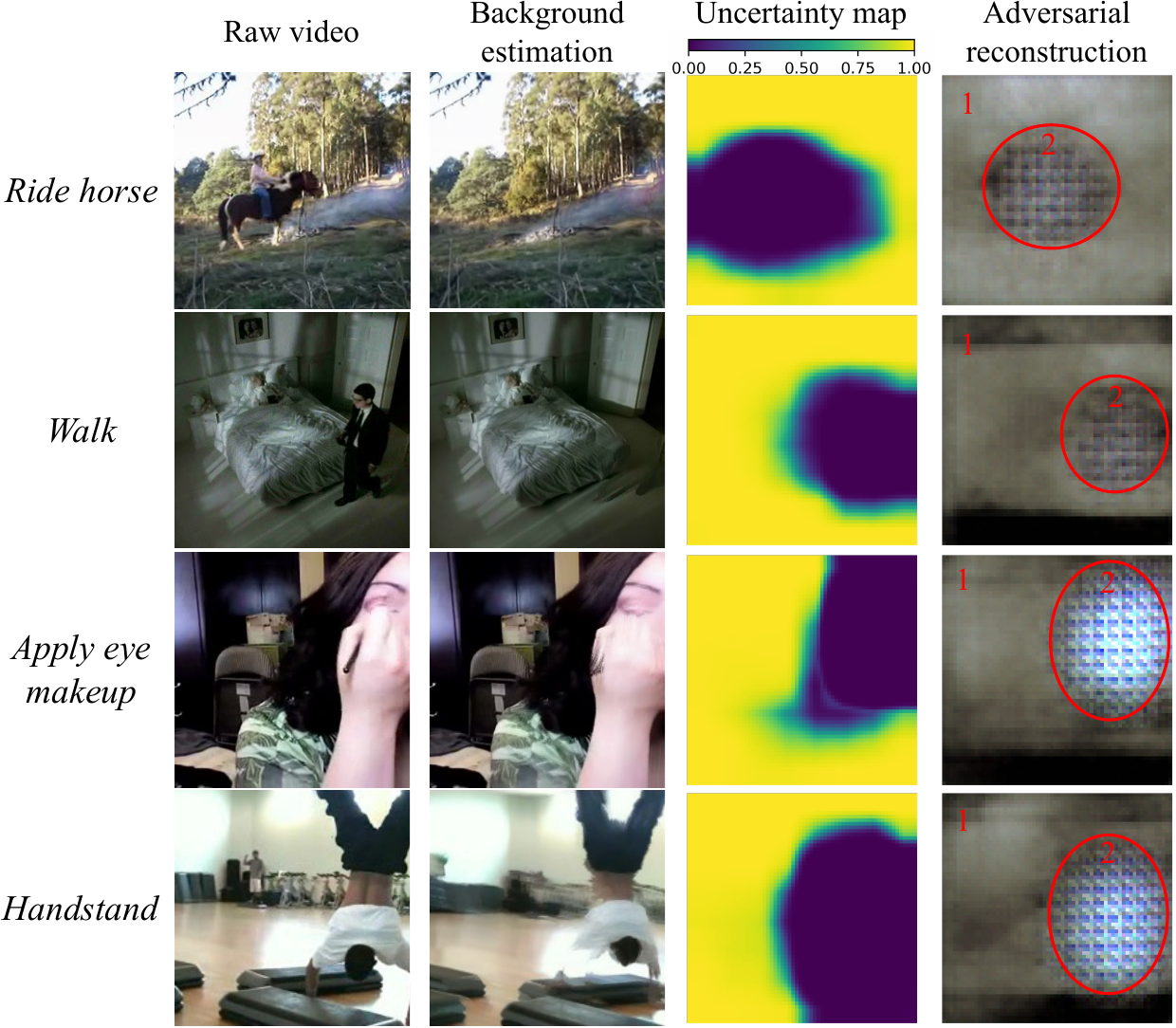}
    \caption{Qualitative results of \reconmod{}. The background estimation
    removes foreground with large motions, and the uncertainty map indicates the
    scene locations (\ie, yellow regions with high uncertainty). \reconmod{}
    reduces the scene information within features extracted by the backbone,
    leading to blurry surroundings (region 1) shown in the $4$-th column. As low
    reconstruction weight is applied on the action foreground (\ie{},
    $u'_{i,j,t}=0$ in~\cref{eq:l_recon}), it is neglected during reconstruction,
    leading to Gaussian-noisy-like reconstruction results (region 2).}
    \label{fig:qualitative-result-on-adv-recon}
\end{figure}

\noindent\textbf{Adversarial scene reconstruction.}
\cref{tab:abaltion-on-adversarial-reconstruction} analyzes the effect of
different designs in \reconmod{}.
First, we observe that simply adversarially reconstructing the raw video has minor improvement on the performance, as such training encourages the backbone to
remove not only static scene information but also foreground motion information.
Subsequently, our background estimation and uncertainty-weighted reconstruction
individually improve the performance, and the best performance is achieved by
combining both.
Additional qualitative results of AdRecon are provided in
\cref{fig:qualitative-result-on-adv-recon}.

\begin{table}[t]
    \centering
    \resizebox{\linewidth}{!}{%
    \begin{tabular}{cc|c|c|c|c}
        \hline 
        $\mathcal{{L}_{\text{{s\_cls}}}}$ & $\mathcal{{L}_{\text{{s\_guide}}}}$ & \multirow{1}{*}{AUC $\uparrow$} & \multirow{1}{*}{FAR@95 $\downarrow$} & \multirow{1}{*}{TPR@10 $\uparrow$} & \multirow{1}{*}{Open maF1 $\uparrow$}\tabularnewline
        \hline 
        - & - & 91.73 & 28.84 & 78.96 & $68.55\pm0.34$\tabularnewline
        \checkmark & - & 92.26 & 29.32 & 82.77 & $71.46\pm0.32$\tabularnewline
        \checkmark & \checkmark & \textbf{93.58} & \textbf{26.43} & \textbf{83.16} & $\mathbf{72.16\pm0.30}$\tabularnewline
        \hline 
    \end{tabular}%
    }
    \caption{Ablation study on the adaptive adversarial scene classification on
    UCF101~\cite{soomro2012ucf101} + MiTv2~\cite{monfortmoments} datasets.}
    \label{tab:abaltion-on-adversarial-scene-recognition}
\end{table}

\noindent\textbf{Adaptive adversarial scene classification.}
\cref{tab:abaltion-on-adversarial-scene-recognition} shows the effectiveness
of \sclsmod{}.
$\mathcal{L}_{\text{s\_cls}}$ improves the OSAR performance as it encourages the
backbone to learn scene-invariant features.
Our proposed uncertainty-guidance loss $\mathcal{L}_{\text{s\_guide}}$ further
improves the performance, demonstrating that the uncertainty map implicitly locates the foreground and guides adversarial scene classification learning.

\section{Conclusion}
\label{sec:conclusion}

In this paper, we propose \abbrmethod{} to mitigate scene bias in OSAR.
Specifically, we spot two typical scenarios where current OSAR methods fail, and
emprically show the scene bias for existing 
methods.
Our \abbrmethod{} features an adversarial scene reconstruction module and an
adaptive adversarial scene classification module.
The former reduces the scene information in the extracted feature to disturb
video scene reconstruction.
The latter learns scene-invariant action features by preventing video scene
classification with a focus on the action foreground.
Our \abbrmethod{} exhibits lower scene bias while achieving state-of-the-art
OSAR performance.

\section*{Acknowledgements}

This work is supported in part by the Defense Advanced Research Projects Agency (DARPA) under Contract No.~HR001120C0124. Any opinions, findings and conclusions or recommendations expressed in this material are those of the author(s) and do not necessarily reflect the views of the Defense Advanced Research Projects Agency (DARPA).

\appendix

\section{Additional experimental results}
\label{sec:exp}

\begin{figure*}[!htb]
    \centering
    \captionsetup[subfloat]{font=small}
    \subfloat[Analysis on the \myblue{known} action in \myred{unfamiliar} scene
    scenario.\label{subfig:tpn-scene-bias-close}]{\includegraphics[width=0.485\linewidth]{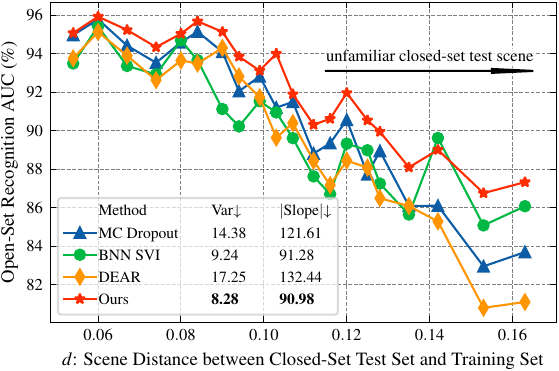}}
    \hfill
    \captionsetup[subfloat]{font=small}
    \subfloat[Analysis on the \myred{unknown} action in \myblue{familiar} scene
    scenario.\label{subfig:tpn-scene-bias-open}]{\includegraphics[width=0.485\linewidth]{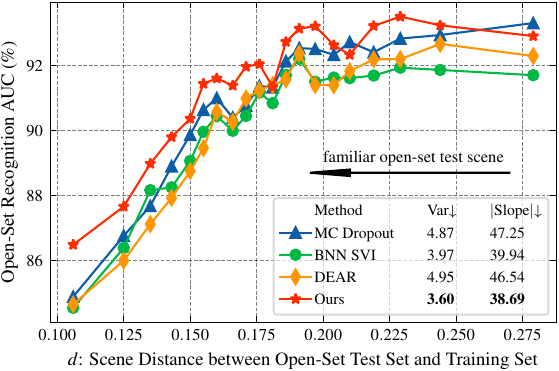}}
    \caption{Quantitative scene bias analysis using
    UCF101~\cite{soomro2012ucf101} as known and MiTv2~\cite{monfortmoments} as
    unknown with the \emph{TPN} backbone~\cite{yang2020temporal}.}
    \label{fig:tpn-scene-bias}
\end{figure*}

\begin{figure*}[!htb]
    \centering
    \captionsetup[subfloat]{font=small}
    \subfloat[Analysis on the \myblue{known} action in \myred{unfamiliar} scene
    scenario.\label{subfig:tsm-scene-bias-close}]{\includegraphics[width=0.485\linewidth]{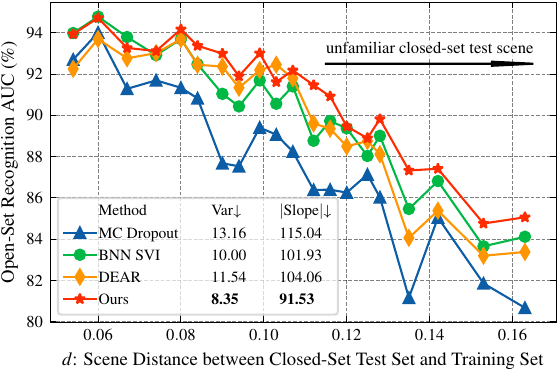}}
    \hfill
    \captionsetup[subfloat]{font=small}
    \subfloat[Analysis on the \myred{unknown} action in \myblue{familiar} scene
    scenario.\label{subfig:tsm-scene-bias-open}]{\includegraphics[width=0.485\linewidth]{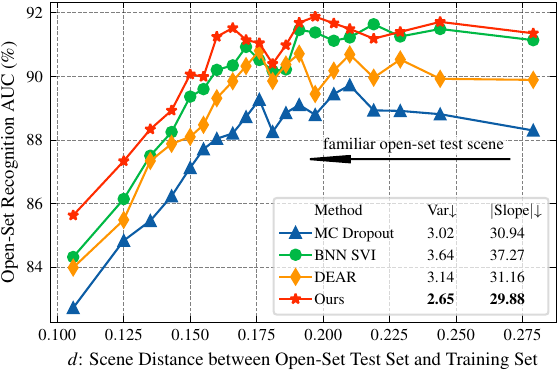}}
    \caption{Quantitative scene bias analysis using
    UCF101~\cite{soomro2012ucf101} as known and MiTv2~\cite{monfortmoments} as
    unknown with the \emph{TSM} backbone~\cite{lin2019tsm}.}
    \label{fig:tsm-scene-bias}
\end{figure*}

\begin{figure*}[!htb]
    \centering
    \captionsetup[subfloat]{font=small}
    \subfloat[Analysis on the \myblue{known} action in \myred{unfamiliar} scene
    scenario.\label{subfig:sf-scene-bias-close}]{\includegraphics[width=0.485\linewidth]{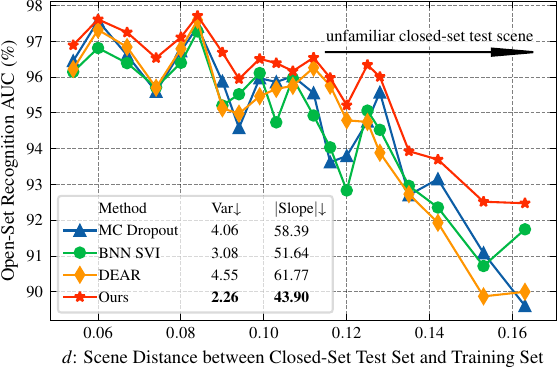}}
    \hfill
    \captionsetup[subfloat]{font=small}
    \subfloat[Analysis on the \myred{unknown} action in \myblue{familiar} scene
    scenario.\label{subfig:sf-scene-bias-open}]{\includegraphics[width=0.485\linewidth]{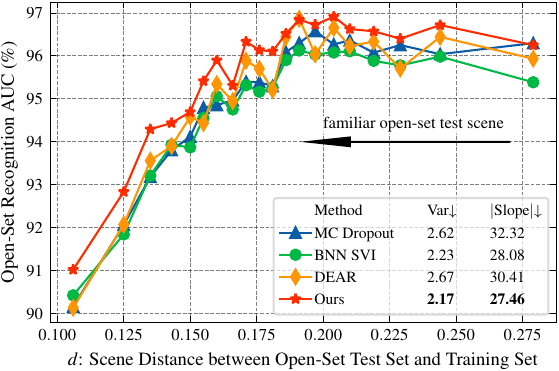}}
    \caption{Quantitative scene bias analysis using
    UCF101~\cite{soomro2012ucf101} as known and MiTv2~\cite{monfortmoments} as
    unknown with the \emph{SlowFast} backbone~\cite{feichtenhofer2019slowfast}.}
    \label{fig:sf-scene-bias}
\end{figure*}

\subsection{Comparison with the state-of-the-art}
In the main paper, we have shown that our \abbrmethod{} outperforms previous
methods in terms of lower scene bias and higher OSAR performance with the I3D
backbone~\cite{carreira2017quo}.
To validate that the superiority of \abbrmethod{} is not tied to a specific
backbone, we carry out experiments with different backbones, \ie{},
SlowFast~\cite{feichtenhofer2019slowfast}, TSM~\cite{lin2019tsm}, and
TPN~\cite{yang2020temporal}.
Furthermore, to analyze how the scene distance affects the overall OSAR and closed-set classification performance,
we carry out the scene bias analysis
experiments using Open maF1 as the metric.

\begin{table*}[t!]
    \centering
    \resizebox{\linewidth}{!}{%
    \begin{tabular}{c|c|c|c|c|c|c|c|c|c|c}
        \hline 
        \multirow{2}{*}{Backbone} & \multirow{2}{*}{Methods} & \multicolumn{4}{c|}{UCF101~\cite{soomro2012ucf101}+MiTv2~\cite{monfortmoments}} & \multicolumn{4}{c|}{UCF101~\cite{soomro2012ucf101}+HMDB51~\cite{kuehne2011hmdb}} & Closed-set\tabularnewline
        \cline{3-10} \cline{4-10} \cline{5-10} \cline{6-10} \cline{7-10} \cline{8-10} \cline{9-10} \cline{10-10} 
         &  & AUC $\uparrow$ & FAR@95 $\downarrow$ & TPR@10 $\uparrow$ & Open maF1 $\uparrow$ & AUC $\uparrow$ & FAR@95 $\downarrow$ & TPR@10 $\uparrow$ & Open maF1 $\uparrow$ & Accuracy\tabularnewline
        \hline 
        \multirow{6}{*}{TPN~\cite{yang2020temporal}} & SoftMax & 43.36 & 97.82 & 8.89 & $55.01\pm0.32$ & 44.92 & 97.33 & 6.42 & $72.31\pm0.12$ & 92.00\tabularnewline
         & OpenMax~\cite{bendale2016towards} & 60.02 & 73.93 & 23.02 & $65.31\pm0.19$ & 62.65 & 64.23 & 19.30 & $65.32\pm0.12$ & 55.37\tabularnewline
         & MC Dropout~\cite{gal2016dropout} & 90.86 & 32.59 & 72.51 & $71.96\pm0.19$ & 84.89 & 64.76 & 57.19 & $77.47\pm0.14$ & 91.28\tabularnewline
         & BNN SVI~\cite{krishnan2018bar} & 90.23 & 32.23 & 67.86 & $69.57\pm0.19$ & 84.93 & 66.82 & 58.82 & $75.38\pm0.15$ & 90.11\tabularnewline
         & DEAR~\cite{bao2021evidential} & 90.31 & 33.67 & 68.32 & $73.57\pm0.19$ & 85.16 & 62.72 & 57.14 & $84.82\pm0.14$ & 92.02\tabularnewline
        \cline{2-11} \cline{3-11} \cline{4-11} \cline{5-11} \cline{6-11} \cline{7-11} \cline{8-11} \cline{9-11} \cline{10-11} \cline{11-11} 
         & \cellcolor{mygray}\abbrmethod{} (Ours) & \textbf{\cellcolor{mygray}91.45} & \textbf{\cellcolor{mygray}30.96} & \textbf{\cellcolor{mygray}74.37} & \textbf{\cellcolor{mygray}}$\mathbf{74.48\pm0.21}$ & \textbf{\cellcolor{mygray}86.67} & \textbf{\cellcolor{mygray}61.02} & \textbf{\cellcolor{mygray}60.62} & \textbf{\cellcolor{mygray}}$\mathbf{85.43\pm0.14}$ & \textbf{\cellcolor{mygray}92.63}\tabularnewline
        \hline 
        \multirow{6}{*}{TSM~\cite{lin2019tsm}} & SoftMax & 46.39 & 94.45 & 9.35 & $54.29\pm0.34$ & 44.58 & 98.44 & 9.32 & $76.29\pm0.19$ & 92.11\tabularnewline
         & OpenMax~\cite{bendale2016towards} & 61.49 & 58.90 & 12.49 & $64.30\pm0.25$ & 60.97 & 63.83 & 10.46 & $64.39\pm0.17$ & 53.48\tabularnewline
         & MC Dropout~\cite{gal2016dropout} & 87.87 & 41.69 & 61.22 & $65.67\pm0.26$ & 84.82 & 63.67 & 63.53 & $75.68\pm0.20$ & 92.15\tabularnewline
         & BNN SVI~\cite{krishnan2018bar} & 89.92 & 40.42 & 72.66 & $65.94\pm0.25$ & 83.28 & 65.96 & 54.31 & $77.63\pm0.19$ & 91.83\tabularnewline
         & DEAR~\cite{bao2021evidential} & 89.12 & 38.98 & 68.07 & $67.33\pm0.36$ & 84.26 & \textbf{57.79} & 62.16 & $86.05\pm0.17$ & 91.94\tabularnewline
        \cline{2-11} \cline{3-11} \cline{4-11} \cline{5-11} \cline{6-11} \cline{7-11} \cline{8-11} \cline{9-11} \cline{10-11} \cline{11-11} 
         & \cellcolor{mygray}\abbrmethod{} (Ours) & \textbf{\cellcolor{mygray}90.47} & \textbf{\cellcolor{mygray}37.17} & \textbf{\cellcolor{mygray}69.69} & \textbf{\cellcolor{mygray}}$\mathbf{69.33\pm0.21}$ & \textbf{\cellcolor{mygray}85.96} & \textbf{\cellcolor{mygray}}60.62 & \textbf{\cellcolor{mygray}65.98} & \textbf{\cellcolor{mygray}}$\mathbf{87.67\pm0.17}$ & \textbf{\cellcolor{mygray}92.49}\tabularnewline
        \hline 
        \multirow{6}{*}{SlowFast~\cite{feichtenhofer2019slowfast}} & SoftMax & 56.02 & 89.33 & 15.54 & $61.12\pm0.26$ & 55.39 & 91.58 & 20.57 & $75.02\pm0.15$ & 96.17\tabularnewline
         & OpenMax~\cite{bendale2016towards} & 68.49 & 39.38 & 10.48 & $69.74\pm0.17$ & 67.00 & 77.54 & 25.35 & $66.46\pm0.16$ & 60.33\tabularnewline
         & MC Dropout~\cite{gal2016dropout} & 95.01 & \textbf{18.21} & 88.99 & $71.12\pm0.16$ & 89.52 & 53.95 & 75.82 & $73.35\pm0.15$ & 96.24\tabularnewline
         & BNN SVI~\cite{krishnan2018bar} & 94.83 & 20.51 & 87.37 & $68.92\pm0.19$ & 88.68 & 60.88 & 74.05 & $71.14\pm0.16$ & 96.01\tabularnewline
         & DEAR~\cite{bao2021evidential} & 95.12 & 20.35 & 87.63 & $75.51\pm0.17$ & 89.33 & 58.78 & 75.95 & $89.71\pm0.17$ & \textbf{96.56}\tabularnewline
        \cline{2-11} \cline{3-11} \cline{4-11} \cline{5-11} \cline{6-11} \cline{7-11} \cline{8-11} \cline{9-11} \cline{10-11} \cline{11-11} 
         & \cellcolor{mygray}\abbrmethod{} (Ours) & \textbf{\cellcolor{mygray}95.72} & \cellcolor{mygray}18.84 & \textbf{\cellcolor{mygray}90.68} & \cellcolor{mygray}$\mathbf{76.47\pm0.14}$ & \textbf{\cellcolor{mygray}90.72} & \textbf{\cellcolor{mygray}52.32} & \textbf{\cellcolor{mygray}76.93} & \textbf{\cellcolor{mygray}$\mathbf{90.64\pm0.19}$} & \cellcolor{mygray}96.53\tabularnewline
        \hline 
    \end{tabular}%
    }
    \caption{Comparison with state-of-the-art methods with different backbones.
    All methods are trained on UCF101~\cite{soomro2012ucf101}, and evaluated on
    two different open sets where unknown samples are from
    HMDB51~\cite{kuehne2011hmdb} and MiTv2~\cite{monfortmoments}, respectively.}
    \label{tab:comparison-with-sota-supp}
\end{table*}

\begin{figure*}[!h]
    \centering
    \captionsetup[subfloat]{font=small}
    \subfloat[Analysis on the \myblue{known} action in \myred{unfamiliar} scene
    scenario.\label{subfig:open-maf1-scene-bias-close}]{\includegraphics[width=0.485\linewidth]{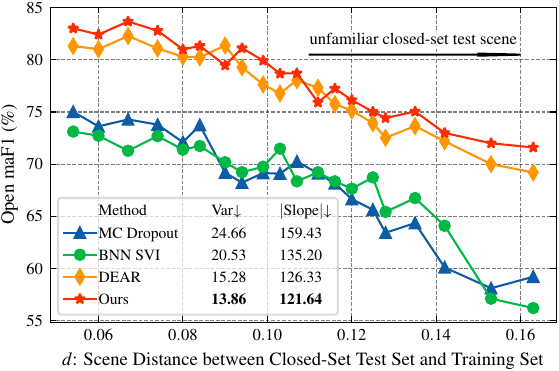}}
    \hfill
    \captionsetup[subfloat]{font=small}
    \subfloat[Analysis on the \myred{unknown} action in \myblue{familiar} scene
    scenario.\label{subfig:open-maf1-scene-bias-open}]{\includegraphics[width=0.485\linewidth]{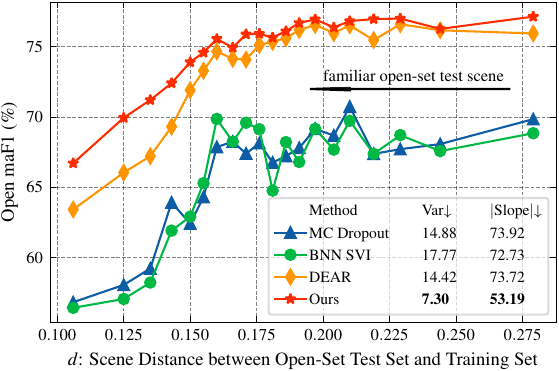}}
    \caption{Quantitative scene bias analysis using \emph{Open maF1}, which
    combines the OSAR and closed-set action recognition performances. The
    experiments are carried out with the I3D backbone~\cite{carreira2017quo},
    using UCF101~\cite{soomro2012ucf101} as known and
    MiTv2~\cite{monfortmoments} as unknown. Our \abbrmethod{} is least affected
    by the scene.}
    \label{fig:open-maf1}
\end{figure*}

\noindent\textbf{Scene bias analysis with different backbones.}
\cref{fig:tpn-scene-bias}, \cref{fig:tsm-scene-bias}, and
\cref{fig:sf-scene-bias} shows the scene bias analysis experiments with
TPN~\cite{yang2020temporal}, TSM~\cite{lin2019tsm}, and
SlowFast~\cite{feichtenhofer2019slowfast} backbones, respectively.
We make the following observations.
(1)~All figures show the same tendency: known actions with unfamiliar scenes
(right part of the left figures) and unknown actions with familiar scenes (left
part of the right figures) are hard to recognize.
This conclusion holds for all backbones, indicating that it is a general problem
rather than a backbone-specific problem.
(2)~Our \abbrmethod{} achieves lower scene bias in both scenarios with all
backbones, showing the generalization ability and debias ability of our method.
Especially, our \abbrmethod{} achieves better OSAR performance when the
closed-set testing set exhibits dissimilar scene to the training set (\ie{}, the
right part of \cref{subfig:tpn-scene-bias-close},
\cref{subfig:tsm-scene-bias-close}, and \cref{subfig:sf-scene-bias-close}), and
when the open-set testing set exhibits similar scene to the training set (\ie{},
the left part of \cref{subfig:tpn-scene-bias-open},
\cref{subfig:tsm-scene-bias-open}, and \cref{subfig:sf-scene-bias-open}).
Such a performance advantage shows that our method successfully avoids the
misleading of scene information, and further demonstrates its debias ability.
We note that this ability is critical when the testing environment is different
from the training environment.

\noindent\textbf{OSAR performance comparison with different backbones.}
\cref{tab:comparison-with-sota-supp} lists the performance comparison with previous
OSAR methods in different backbones.
The results show that our \abbrmethod{} achieves state-of-the-art OSAR
performance and outperforms all previous methods in terms of AUC and open macro
F1 with all backbones, demonstrating the effectiveness of our method.

\noindent\textbf{Scene bias analysis using Open maF1.}
In Fig.~2 of the main paper, we show a strong correlation between OSAR
performance and the scene distance.
We further illustrate how the scene distance affects the overall OSAR and
closed-set classification performance (\ie{}, the C + 1 way classification performance) by conducting the scene bias analysis
using Open maF1 in~\cref{fig:open-maf1}.
The results reveal a similar trend that the scene distance and Open maF1 is highly correlated, and our~\abbrmethod{} achieves the best
performance as well as the lowest scene bias, demonstrating its effectiveness.

\subsection{Ablation study on
UCF101~\cite{soomro2012ucf101}+HMDB51~\cite{kuehne2011hmdb}}

To demonstrate the generalization ability of our proposed \reconmod{} and
\sclsmod{}, we further show the results of the ablation study in the
UCF101~\cite{soomro2012ucf101}+HMDB51~\cite{kuehne2011hmdb} testing set in
\cref{tab:ablation-on-adversarial-reconstruction} and
\cref{tab:ablation-on-adversarial-scene-recognition}, respectively.
The results reveal that all designs in both modules contribute to the final
performance, which aligns with the conclusion made in the main paper,
demonstrating the generalization ability of our method.

\begin{table}[t]
    \centering
    \resizebox{\linewidth}{!}{%
    \begin{tabular}{ccc|c|c|c|c}
        \hline 
        \multirow{2}{*}{\reconmod{}} & Bg. & Unc. & \multirow{2}{*}{AUC $\uparrow$} & \multirow{2}{*}{FAR@95 $\downarrow$} & \multirow{2}{*}{TPR@10 $\uparrow$} & \multirow{2}{*}{Open maF1 $\uparrow$}\tabularnewline
         & Est. & Weight &  &  &  & \tabularnewline
        \hline 
        - & - & - & 85.63 & 78.59 & 68.10 & $87.73\pm0.22$\tabularnewline
        \checkmark & - & - & 85.72 & 82.66 & 71.63 & $86.68\pm0.21$\tabularnewline
        \checkmark & - & \checkmark & 87.17 & 69.82 & 70.46 & $88.08\pm0.20$\tabularnewline
        \checkmark & \checkmark & - & 86.95 & 70.18 & 71.76 & $88.01\pm0.21$\tabularnewline
        \checkmark & \checkmark & \checkmark & \textbf{87.49} & \textbf{69.41} & \textbf{72.31} & $\mathbf{89.52\pm0.21}$\tabularnewline
        \hline 
    \end{tabular}%
    }
    \caption{Ablation study on the adversarial reconstruction on
    UCF101~\cite{soomro2012ucf101} + HMDB51~\cite{kuehne2011hmdb} datasets.}
    \label{tab:ablation-on-adversarial-reconstruction}
\end{table}

\begin{table}[t]
    \centering
    \resizebox{\linewidth}{!}{%
    \begin{tabular}{cc|c|c|c|c}
        \hline 
        $\mathcal{{L}_{\text{{s\_cls}}}}$ & $\mathcal{{L}_{\text{{s\_guide}}}}$ & \multirow{1}{*}{AUC $\uparrow$} & \multirow{1}{*}{FAR@95 $\downarrow$} & \multirow{1}{*}{TPR@10 $\uparrow$} & \multirow{1}{*}{Open maF1 $\uparrow$}\tabularnewline
        \hline 
        - & - & 85.63 & 78.59 & 68.10 & $\mathbf{87.73\pm0.22}$\tabularnewline
        \checkmark & - & 86.87 & 73.42 & 68.48 & $87.42\pm0.23$\tabularnewline
        \checkmark & \checkmark & \textbf{87.22} & \textbf{71.45} & \textbf{69.80} & $87.47\pm0.19$\tabularnewline
        \hline 
    \end{tabular}%
    }
    \caption{Ablation study on the adversarial scene classification on
    UCF101~\cite{soomro2012ucf101} + HMDB51~\cite{kuehne2011hmdb} datasets.}
    \label{tab:ablation-on-adversarial-scene-recognition}
\end{table}

\section{Additional implementation details}
\label{sec:details}

For the TPN backbone~\cite{yang2020temporal}, we follow
DEAR~\cite{bao2021evidential} to use the slow-only version for feature
extraction.
For the TSM backbone~\cite{lin2019tsm}, we use the default setting in
MMAction2~\cite{2020mmaction2} for feature extraciton following
DEAR~\cite{bao2021evidential}.
For the SlowFast backbone~\cite{feichtenhofer2019slowfast}, we interpolate the
extracted spatio-temporal features from the slow and fast pathways to the same
size, and concatenate them in the channel dimension as the final spatio-temporal
feature $\bm{F}$.

We note that our reported results are different from those reported in
DEAR~\cite{bao2021evidential} as they use binarized prediction for the AUC
prediction, which only has one operating point on the ROC curve, while we use the raw
prediction for the AUC computation.

{\small
\bibliographystyle{ieee_fullname}
\bibliography{egbib}

\begin{thebibliography}{10}\itemsep=-1pt

\bibitem{amini2020deep}
Alexander Amini, Wilko Schwarting, Ava Soleimany, and Daniela Rus.
\newblock Deep evidential regression.
\newblock {\em NeurIPS}, 33:14927--14937, 2020.

\bibitem{bahng2020learning}
Hyojin Bahng, Sanghyuk Chun, Sangdoo Yun, Jaegul Choo, and Seong~Joon Oh.
\newblock Learning de-biased representations with biased representations.
\newblock In {\em ICML}, pages 528--539, 2020.

\bibitem{bao2021evidential}
Wentao Bao, Qi Yu, and Yu Kong.
\newblock Evidential deep learning for open set action recognition.
\newblock In {\em ICCV}, pages 13349--13358, 2021.

\bibitem{bendale2015towards}
Abhijit Bendale and Terrance Boult.
\newblock Towards open world recognition.
\newblock In {\em CVPR}, pages 1893--1902, 2015.

\bibitem{bendale2016towards}
Abhijit Bendale and Terrance~E Boult.
\newblock Towards open set deep networks.
\newblock In {\em CVPR}, pages 1563--1572, 2016.

\bibitem{blundell2015weight}
Charles Blundell, Julien Cornebise, Koray Kavukcuoglu, and Daan Wierstra.
\newblock Weight uncertainty in neural network.
\newblock In {\em ICML}, pages 1613--1622, 2015.

\bibitem{bolukbasi2016man}
Tolga Bolukbasi, Kai-Wei Chang, James~Y Zou, Venkatesh Saligrama, and Adam~T
  Kalai.
\newblock Man is to computer programmer as woman is to homemaker? debiasing
  word embeddings.
\newblock {\em NeurIPS}, 29, 2016.

\bibitem{buolamwini2018gender}
Joy Buolamwini and Timnit Gebru.
\newblock Gender shades: Intersectional accuracy disparities in commercial
  gender classification.
\newblock In {\em Conference on fairness, accountability and transparency},
  pages 77--91, 2018.

\bibitem{busto2018open}
Pau~Panareda Busto, Ahsan Iqbal, and Juergen Gall.
\newblock Open set domain adaptation for image and action recognition.
\newblock {\em IEEE TPAMI}, pages 413--429, 2018.

\bibitem{carreira2017quo}
Joao Carreira and Andrew Zisserman.
\newblock Quo vadis, action recognition? a new model and the kinetics dataset.
\newblock In {\em CVPR}, pages 6299--6308, 2017.

\bibitem{chen2021adversarial}
G Chen, P Peng, X Wang, and Y Tian.
\newblock Adversarial reciprocal points learning for open set recognition.
\newblock {\em IEEE TPAMI}, 2021.

\bibitem{chen2020learning}
Guangyao Chen, Limeng Qiao, Yemin Shi, Peixi Peng, Jia Li, Tiejun Huang,
  Shiliang Pu, and Yonghong Tian.
\newblock Learning open set network with discriminative reciprocal points.
\newblock In {\em ECCV}, pages 507--522, 2020.

\bibitem{choi2019can}
Jinwoo Choi, Chen Gao, Joseph~CE Messou, and Jia-Bin Huang.
\newblock Why can't i dance in the mall? learning to mitigate scene bias in
  action recognition.
\newblock In {\em NeurIPS}, 2019.

\bibitem{2020mmaction2}
MMAction2 Contributors.
\newblock Openmmlab's next generation video understanding toolbox and
  benchmark.
\newblock \url{https://github.com/open-mmlab/mmaction2}, 2020.

\bibitem{corbiere2019addressing}
Charles Corbi{\`e}re, Nicolas Thome, Avner Bar-Hen, Matthieu Cord, and Patrick
  P{\'e}rez.
\newblock Addressing failure prediction by learning model confidence.
\newblock {\em NeurIPS}, 32, 2019.

\bibitem{crasto2019mars}
Nieves Crasto, Philippe Weinzaepfel, Karteek Alahari, and Cordelia Schmid.
\newblock Mars: Motion-augmented rgb stream for action recognition.
\newblock In {\em CVPR}, pages 7882--7891, 2019.

\bibitem{dempster1968generalization}
Arthur~P Dempster.
\newblock A generalization of bayesian inference.
\newblock {\em Journal of the Royal Statistical Society: Series B
  (Methodological)}, 30(2):205--232, 1968.

\bibitem{ditria2020opengan}
Luke Ditria, Benjamin~J Meyer, and Tom Drummond.
\newblock Opengan: Open set generative adversarial networks.
\newblock In {\em ACCV}, 2020.

\bibitem{elazar2018adversarial}
Yanai Elazar and Yoav Goldberg.
\newblock Adversarial removal of demographic attributes from text data.
\newblock {\em arXiv preprint arXiv:1808.06640}, 2018.

\bibitem{feichtenhofer2019slowfast}
Christoph Feichtenhofer, Haoqi Fan, Jitendra Malik, and Kaiming He.
\newblock Slowfast networks for video recognition.
\newblock In {\em ICCV}, pages 6202--6211, 2019.

\bibitem{gal2016dropout}
Yarin Gal and Zoubin Ghahramani.
\newblock Dropout as a bayesian approximation: Representing model uncertainty
  in deep learning.
\newblock In {\em ICML}, pages 1050--1059, 2016.

\bibitem{ganin2015unsupervised}
Yaroslav Ganin and Victor Lempitsky.
\newblock Unsupervised domain adaptation by backpropagation.
\newblock In {\em ICML}, pages 1180--1189, 2015.

\bibitem{ge2017generative}
ZongYuan Ge, Sergey Demyanov, Zetao Chen, and Rahil Garnavi.
\newblock Generative openmax for multi-class open set classification.
\newblock {\em arXiv preprint arXiv:1707.07418}, 2017.

\bibitem{geirhos2018imagenet}
Robert Geirhos, Patricia Rubisch, Claudio Michaelis, Matthias Bethge, Felix~A
  Wichmann, and Wieland Brendel.
\newblock Imagenet-trained cnns are biased towards texture; increasing shape
  bias improves accuracy and robustness.
\newblock {\em arXiv preprint arXiv:1811.12231}, 2018.

\bibitem{geng2020recent}
Chuanxing Geng, Sheng-jun Huang, and Songcan Chen.
\newblock Recent advances in open set recognition: A survey.
\newblock {\em IEEE TPAMI}, pages 3614--3631, 2020.

\bibitem{guo2022uncertainty}
Hongji Guo, Zhou Ren, Yi Wu, Gang Hua, and Qiang Ji.
\newblock Uncertainty-based spatial-temporal attention for online action
  detection.
\newblock In {\em ECCV}, pages 69--86. Springer, 2022.

\bibitem{he2016deep}
Kaiming He, Xiangyu Zhang, Shaoqing Ren, and Jian Sun.
\newblock Deep residual learning for image recognition.
\newblock In {\em CVPR}, pages 770--778, 2016.

\bibitem{hendricks2018women}
Lisa~Anne Hendricks, Kaylee Burns, Kate Saenko, Trevor Darrell, and Anna
  Rohrbach.
\newblock Women also snowboard: Overcoming bias in captioning models.
\newblock In {\em ECCV}, pages 771--787, 2018.

\bibitem{hinton2006reducing}
Geoffrey~E Hinton and Ruslan~R Salakhutdinov.
\newblock Reducing the dimensionality of data with neural networks.
\newblock {\em science}, 313(5786):504--507, 2006.

\bibitem{hu20222tad}
Xin Hu, Zhenyu Wu, Hao-Yu Miao, Siqi Fan, Taiyu Long, Zhenyu Hu, Pengcheng Pi,
  Yi Wu, Zhou Ren, Zhangyang Wang, et~al.
\newblock E\^{} 2tad: An energy-efficient tracking-based action detector.
\newblock {\em arXiv preprint arXiv:2204.04416}, 2022.

\bibitem{jain2014multi}
Lalit~P Jain, Walter~J Scheirer, and Terrance~E Boult.
\newblock Multi-class open set recognition using probability of inclusion.
\newblock In {\em ECCV}, pages 393--409, 2014.

\bibitem{josang2016subjective}
Audun J{\o}sang.
\newblock {\em Subjective logic}, volume~3.
\newblock 2016.

\bibitem{kim2019learning}
Byungju Kim, Hyunwoo Kim, Kyungsu Kim, Sungjin Kim, and Junmo Kim.
\newblock Learning not to learn: Training deep neural networks with biased
  data.
\newblock In {\em CVPR}, pages 9012--9020, 2019.

\bibitem{kong2021opengan}
Shu Kong and Deva Ramanan.
\newblock Opengan: Open-set recognition via open data generation.
\newblock In {\em ICCV}, pages 813--822, 2021.

\bibitem{kornblith2019similarity}
Simon Kornblith, Mohammad Norouzi, Honglak Lee, and Geoffrey Hinton.
\newblock Similarity of neural network representations revisited.
\newblock In {\em ICML}, pages 3519--3529, 2019.

\bibitem{krishnan2018bar}
Ranganath Krishnan, Mahesh Subedar, and Omesh Tickoo.
\newblock Bar: Bayesian activity recognition using variational inference.
\newblock {\em NeurIPS Workshop}, 2018.

\bibitem{krishnan2020specifying}
Ranganath Krishnan, Mahesh Subedar, and Omesh Tickoo.
\newblock Specifying weight priors in bayesian deep neural networks with
  empirical bayes.
\newblock In {\em AAAI}, pages 4477--4484, 2020.

\bibitem{kuehne2011hmdb}
Hildegard Kuehne, Hueihan Jhuang, Est{\'\i}baliz Garrote, Tomaso Poggio, and
  Thomas Serre.
\newblock Hmdb: a large video database for human motion recognition.
\newblock In {\em ICCV}, pages 2556--2563, 2011.

\bibitem{li2022trustworthy}
Bolian Li, Zongbo Han, Haining Li, Huazhu Fu, and Changqing Zhang.
\newblock Trustworthy long-tailed classification.
\newblock In {\em CVPR}, pages 6970--6979, 2022.

\bibitem{li2018resound}
Yingwei Li, Yi Li, and Nuno Vasconcelos.
\newblock Resound: Towards action recognition without representation bias.
\newblock In {\em ECCV}, pages 513--528, 2018.

\bibitem{lin2019tsm}
Ji Lin, Chuang Gan, and Song Han.
\newblock Tsm: Temporal shift module for efficient video understanding.
\newblock In {\em ICCV}, pages 7083--7093, 2019.

\bibitem{liu2016scene}
Wei Liu, Yuanzheng Cai, Miaohui Zhang, Hui Li, and Hejin Gu.
\newblock Scene background estimation based on temporal median filter with
  gaussian filtering.
\newblock In {\em ICPR}, pages 132--136, 2016.

\bibitem{mandal2020motionrec}
Murari Mandal, Lav~Kush Kumar, Mahipal~Singh Saran, et~al.
\newblock Motionrec: A unified deep framework for moving object recognition.
\newblock In {\em WACV}, pages 2734--2743, 2020.

\bibitem{mo2021object}
Sangwoo Mo, Hyunwoo Kang, Kihyuk Sohn, Chun-Liang Li, and Jinwoo Shin.
\newblock Object-aware contrastive learning for debiased scene representation.
\newblock {\em NeurIPS}, 34:12251--12264, 2021.

\bibitem{monfortmoments}
Mathew Monfort, Alex Andonian, Bolei Zhou, Kandan Ramakrishnan, Sarah~Adel
  Bargal, Tom Yan, Lisa Brown, Quanfu Fan, Dan Gutfruend, Carl Vondrick, et~al.
\newblock Moments in time dataset: one million videos for event understanding.
\newblock {\em IEEE TPAMI}, pages 1--8, 2019.

\bibitem{moon2020confidence}
Jooyoung Moon, Jihyo Kim, Younghak Shin, and Sangheum Hwang.
\newblock Confidence-aware learning for deep neural networks.
\newblock In {\em ICML}, pages 7034--7044, 2020.

\bibitem{oza2019c2ae}
Poojan Oza and Vishal~M Patel.
\newblock C2ae: Class conditioned auto-encoder for open-set recognition.
\newblock In {\em Proceedings of the IEEE/CVF Conference on Computer Vision and
  Pattern Recognition}, pages 2307--2316, 2019.

\bibitem{pytorch}
Adam Paszke, Sam Gross, Francisco Massa, Adam Lerer, James Bradbury, Gregory
  Chanan, Trevor Killeen, Zeming Lin, Natalia Gimelshein, Luca Antiga, Alban
  Desmaison, Andreas Kopf, Edward Yang, Zachary DeVito, Martin Raison, Alykhan
  Tejani, Sasank Chilamkurthy, Benoit Steiner, Lu Fang, Junjie Bai, and Soumith
  Chintala.
\newblock Pytorch: An imperative style, high-performance deep learning library.
\newblock In {\em NeurIPS}, pages 8024--8035. 2019.

\bibitem{pathak2014fully}
Deepak Pathak, Evan Shelhamer, Jonathan Long, and Trevor Darrell.
\newblock Fully convolutional multi-class multiple instance learning.
\newblock {\em arXiv preprint arXiv:1412.7144}, 2014.

\bibitem{perera2020generative}
Pramuditha Perera, Vlad~I Morariu, Rajiv Jain, Varun Manjunatha, Curtis
  Wigington, Vicente Ordonez, and Vishal~M Patel.
\newblock Generative-discriminative feature representations for open-set
  recognition.
\newblock In {\em CVPR}, pages 11814--11823, 2020.

\bibitem{piccardi2004background}
Massimo Piccardi.
\newblock Background subtraction techniques: a review.
\newblock In {\em 2004 IEEE international conference on systems, man and
  cybernetics (IEEE Cat. No. 04CH37583)}, volume~4, pages 3099--3104, 2004.

\bibitem{piergiovanni2019representation}
AJ Piergiovanni and Michael~S Ryoo.
\newblock Representation flow for action recognition.
\newblock In {\em CVPR}, pages 9945--9953, 2019.

\bibitem{pinheiro2015image}
Pedro~O Pinheiro and Ronan Collobert.
\newblock From image-level to pixel-level labeling with convolutional networks.
\newblock In {\em CVPR}, pages 1713--1721, 2015.

\bibitem{ramachandra2020survey}
Bharathkumar Ramachandra, Michael Jones, and Ranga~Raju Vatsavai.
\newblock A survey of single-scene video anomaly detection.
\newblock {\em IEEE TPAMI}, 2020.

\bibitem{sahoo2021contrast}
Aadarsh Sahoo, Rutav Shah, Rameswar Panda, Kate Saenko, and Abir Das.
\newblock Contrast and mix: Temporal contrastive video domain adaptation with
  background mixing.
\newblock {\em NeurIPS}, 34:23386--23400, 2021.

\bibitem{scheirer2012toward}
Walter~J Scheirer, Anderson de Rezende~Rocha, Archana Sapkota, and Terrance~E
  Boult.
\newblock Toward open set recognition.
\newblock {\em IEEE TPAMI}, pages 1757--1772, 2012.

\bibitem{scheirer2014probability}
Walter~J Scheirer, Lalit~P Jain, and Terrance~E Boult.
\newblock Probability models for open set recognition.
\newblock {\em IEEE TPAMI}, pages 2317--2324, 2014.

\bibitem{sensoy2018evidential}
Murat Sensoy, Lance Kaplan, and Melih Kandemir.
\newblock Evidential deep learning to quantify classification uncertainty.
\newblock {\em NeurIPS}, 31, 2018.

\bibitem{shou2019dmc}
Zheng Shou, Xudong Lin, Yannis Kalantidis, Laura Sevilla-Lara, Marcus Rohrbach,
  Shih-Fu Chang, and Zhicheng Yan.
\newblock Dmc-net: Generating discriminative motion cues for fast compressed
  video action recognition.
\newblock In {\em CVPR}, pages 1268--1277, 2019.

\bibitem{shu2017doc}
Lei Shu, Hu Xu, and Bing Liu.
\newblock Doc: Deep open classification of text documents.
\newblock {\em arXiv preprint arXiv:1709.08716}, 2017.

\bibitem{shu2018odn}
Yu Shu, Yemin Shi, Yaowei Wang, Yixiong Zou, Qingsheng Yuan, and Yonghong Tian.
\newblock Odn: Opening the deep network for open-set action recognition.
\newblock In {\em ICME}, pages 1--6, 2018.

\bibitem{simonyan2014two}
Karen Simonyan and Andrew Zisserman.
\newblock Two-stream convolutional networks for action recognition in videos.
\newblock {\em NeurIPS}, 27, 2014.

\bibitem{soomro2012ucf101}
Khurram Soomro, Amir~Roshan Zamir, and Mubarak Shah.
\newblock Ucf101: A dataset of 101 human actions classes from videos in the
  wild.
\newblock {\em arXiv preprint arXiv:1212.0402}, 2012.

\bibitem{subedar2019uncertainty}
Mahesh Subedar, Ranganath Krishnan, Paulo~Lopez Meyer, Omesh Tickoo, and
  Jonathan Huang.
\newblock Uncertainty-aware audiovisual activity recognition using deep
  bayesian variational inference.
\newblock In {\em ICCV}, pages 6301--6310, 2019.

\bibitem{sun2020conditional}
Xin Sun, Zhenning Yang, Chi Zhang, Keck-Voon Ling, and Guohao Peng.
\newblock Conditional gaussian distribution learning for open set recognition.
\newblock In {\em CVPR}, pages 13480--13489, 2020.

\bibitem{tezcan2020bsuv}
Ozan Tezcan, Prakash Ishwar, and Janusz Konrad.
\newblock Bsuv-net: A fully-convolutional neural network for background
  subtraction of unseen videos.
\newblock In {\em WACV}, pages 2774--2783, 2020.

\bibitem{van2020uncertainty}
Joost Van~Amersfoort, Lewis Smith, Yee~Whye Teh, and Yarin Gal.
\newblock Uncertainty estimation using a single deep deterministic neural
  network.
\newblock In {\em ICML}, pages 9690--9700, 2020.

\bibitem{wang2019hallucinating}
Lei Wang, Piotr Koniusz, and Du~Q Huynh.
\newblock Hallucinating idt descriptors and i3d optical flow features for
  action recognition with cnns.
\newblock In {\em ICCV}, pages 8698--8708, 2019.

\bibitem{wang2018temporal}
Limin Wang, Yuanjun Xiong, Zhe Wang, Yu Qiao, Dahua Lin, Xiaoou Tang, and Luc
  Van~Gool.
\newblock Temporal segment networks for action recognition in videos.
\newblock {\em IEEE TPAMI}, (11):2740--2755, 2018.

\bibitem{wang2018non}
Xiaolong Wang, Ross Girshick, Abhinav Gupta, and Kaiming He.
\newblock Non-local neural networks.
\newblock In {\em CVPR}, pages 7794--7803, 2018.

\bibitem{weinzaepfel2021mimetics}
Philippe Weinzaepfel and Gr{\'e}gory Rogez.
\newblock Mimetics: Towards understanding human actions out of context.
\newblock {\em IJCV}, 129(5):1675--1690, 2021.

\bibitem{wu2022txvad}
Zhenyu Wu, Zhou Ren, Yi Wu, Zhangyang Wang, and Gang Hua.
\newblock Txvad: Improved video action detection by transformers.
\newblock In {\em ACM MM}, pages 4605--4613, 2022.

\bibitem{wu2019delving}
Zhenyu Wu, Karthik Suresh, Priya Narayanan, Hongyu Xu, Heesung Kwon, and
  Zhangyang Wang.
\newblock Delving into robust object detection from unmanned aerial vehicles: A
  deep nuisance disentanglement approach.
\newblock In {\em ICCV}, pages 1201--1210, 2019.

\bibitem{wu2020privacy}
Zhenyu Wu, Haotao Wang, Zhaowen Wang, Hailin Jin, and Zhangyang Wang.
\newblock Privacy-preserving deep action recognition: An adversarial learning
  framework and a new dataset.
\newblock {\em IEEE TPAMI}, 2020.

\bibitem{wu2018towards}
Zhenyu Wu, Zhangyang Wang, Zhaowen Wang, and Hailin Jin.
\newblock Towards privacy-preserving visual recognition via adversarial
  training: A pilot study.
\newblock In {\em ECCV}, pages 606--624, 2018.

\bibitem{xie2017controllable}
Qizhe Xie, Zihang Dai, Yulun Du, Eduard Hovy, and Graham Neubig.
\newblock Controllable invariance through adversarial feature learning.
\newblock {\em NeurIPS}, 30, 2017.

\bibitem{yang2020temporal}
Ceyuan Yang, Yinghao Xu, Jianping Shi, Bo Dai, and Bolei Zhou.
\newblock Temporal pyramid network for action recognition.
\newblock In {\em CVPR}, pages 591--600, 2020.

\bibitem{yoshihashi2019classification}
Ryota Yoshihashi, Wen Shao, Rei Kawakami, Shaodi You, Makoto Iida, and Takeshi
  Naemura.
\newblock Classification-reconstruction learning for open-set recognition.
\newblock In {\em CVPR}, pages 4016--4025, 2019.

\bibitem{yu2019temporal}
Tan Yu, Zhou Ren, Yuncheng Li, Enxu Yan, Ning Xu, and Junsong Yuan.
\newblock Temporal structure mining for weakly supervised action detection.
\newblock In {\em ICCV}, pages 5522--5531, 2019.

\bibitem{yue2021counterfactual}
Zhongqi Yue, Tan Wang, Qianru Sun, Xian-Sheng Hua, and Hanwang Zhang.
\newblock Counterfactual zero-shot and open-set visual recognition.
\newblock In {\em CVPR}, pages 15404--15414, 2021.

\bibitem{zhai2020two}
Yuanhao Zhai, Le Wang, Wei Tang, Qilin Zhang, Junsong Yuan, and Gang Hua.
\newblock Two-stream consensus network for weakly-supervised temporal action
  localization.
\newblock In {\em ECCV}, pages 37--54, 2020.

\bibitem{zhai2022adaptive}
Yuanhao Zhai, Le Wang, Wei Tang, Qilin Zhang, Nanning Zheng, David Doermann,
  Junsong Yuan, and Gang Hua.
\newblock Adaptive two-stream consensus network for weakly-supervised temporal
  action localization.
\newblock {\em IEEE TPAMI}, 45(4):4136--4151, 2022.

\bibitem{zhai2021action}
Yuanhao Zhai, Le Wang, Wei Tang, Qilin Zhang, Nanning Zheng, and Gang Hua.
\newblock Action coherence network for weakly-supervised temporal action
  localization.
\newblock {\em IEEE TMM}, 24:1857--1870, 2021.

\bibitem{zhao2017men}
Jieyu Zhao, Tianlu Wang, Mark Yatskar, Vicente Ordonez, and Kai-Wei Chang.
\newblock Men also like shopping: Reducing gender bias amplification using
  corpus-level constraints.
\newblock {\em arXiv preprint arXiv:1707.09457}, 2017.

\bibitem{zhao2018learning}
Jieyu Zhao, Yichao Zhou, Zeyu Li, Wei Wang, and Kai-Wei Chang.
\newblock Learning gender-neutral word embeddings.
\newblock {\em arXiv preprint arXiv:1809.01496}, 2018.

\bibitem{zhao2018recognize}
Yue Zhao, Yuanjun Xiong, and Dahua Lin.
\newblock Recognize actions by disentangling components of dynamics.
\newblock In {\em CVPR}, pages 6566--6575, 2018.

\bibitem{zhou2016learning}
Bolei Zhou, Aditya Khosla, Agata Lapedriza, Aude Oliva, and Antonio Torralba.
\newblock Learning deep features for discriminative localization.
\newblock In {\em CVPR}, pages 2921--2929, 2016.

\bibitem{zhou2017places}
Bolei Zhou, Agata Lapedriza, Aditya Khosla, Aude Oliva, and Antonio Torralba.
\newblock Places: A 10 million image database for scene recognition.
\newblock {\em IEEE TPAMI}, 2017.

\bibitem{zhou2021learning}
Da-Wei Zhou, Han-Jia Ye, and De-Chuan Zhan.
\newblock Learning placeholders for open-set recognition.
\newblock In {\em CVPR}, pages 4401--4410, 2021.

\end{thebibliography}
}

\end{document}